\documentclass[final,12pt]{clear2022}

\usepackage{graphicx}
\usepackage[labelfont=bf]{caption}

\usepackage{booktabs}

\usepackage{listings}
\usepackage{fancyvrb}
\fvset{fontsize=\normalsize}

\usepackage[acronym,shortcuts,smallcaps,nowarn,nohypertypes={acronym,notation}]{glossaries}

\lstdefinestyle{mystyle}{
    commentstyle=\color{OliveGreen},
    keywordstyle=\color{BurntOrange},
    numberstyle=\tiny\color{black!60},
    stringstyle=\color{MidnightBlue},
    basicstyle=\ttfamily,
    breakatwhitespace=false,
    breaklines=true,
    captionpos=b,
    keepspaces=true,
    numbers=left,
    numbersep=5pt,
    showspaces=false,
    showstringspaces=false,
    showtabs=false,
    tabsize=2
}
\usepackage[nameinlink]{cleveref}

\usepackage{tikz}
\usetikzlibrary{shapes,decorations,arrows,calc,arrows.meta,fit,positioning}
\tikzset{
    -Latex,auto,node distance =1 cm and 1 cm,semithick,
    state/.style ={circle, draw, minimum width = 0.7 cm},
    detstate/.style ={rectangle, draw, minimum width = 0.7 cm, minimum height = 0.7 cm},
    point/.style = {circle, draw, inner sep=0.04cm,fill,node contents={}},
    bidirected/.style={Latex-Latex,dashed},
    el/.style = {inner sep=2pt, align=left, sloped}
}

\usepackage{wrapfig}
\usepackage{mathtools}

\def\E{\mathop{\mathbb{E}}} 

\def\wp{W^\prime} 
\def\xp{X^\prime} 
\def\zp{Z^\prime}

\newacronym{kl}{kl}{Kullback-Leibler}
\newacronym{ELBO}{elbo}{\emph{evidence lower bound}}
\newacronym{POPELBO}{pop-elbo}{\emph{population evidence lower bound}}

\newacronym{SVI}{svi}{stochastic variational inference}
\newacronym{conc}{conc}{concordance}

\newacronym{svm}{svm}{svm}

\newacronym{BUMPVI}{bump-vi}{bumping variational inference}

\newacronym{GMM}{gmm}{Gaussian mixture model}
\newacronym{LDA}{lda}{latent Dirichlet allocation}

\newacronym{SUTVA}{sutva}{stable unit treatment value assumption}

\newacronym{KSD}{ksd}{{kernelized Stein discrepancy}}
\newacronym{KCC-SD}{kcc-sd}{kernelized complete conditional Stein discrepancy}

\newacronym{OPVI}{opvi}{operator variational inference}
\newacronym{SVGD}{svgd}{Stein variational gradient descent}

\newacronym{vde}{vde}{variational decoupling}
\newacronym{cfn}{cfn}{control-function method}
\newacronym{gcfn}{gcfn}{generalized control-function method}
\newacronym{2sls}{2sls}{two-stage least-squares method}
\newacronym{gmm}{gmm}{generalized method of moments}
\newacronym{iv}{iv}{instrumental variable}
\newacronym{cdf}{cdf}{cumulative distribution function}

\newacronym{x-cal}{x-cal}{explicit calibration}

\newacronym{d-cal}{d-calibration}{distributional calibration}
\newacronym{d-cal-short}{d-cal}{d-cal}

\newacronym{nll}{nll}{negative log likelihood}
\newacronym{ipcw}{ipcw}{inverse probability of censor-weighting}
\newacronym{ip}{ip}{inverse probability}

\newacronym{erm}{erm}{empirical risk minimization}

\newacronym{bs}{bs}{Brier score}

\newacronym{fbs}{fbs}{F Brier score}
\newacronym{gbs}{gbs}{G Brier score}

\newacronym{fbscw}{fbscw}{Weighted F Brier score}
\newacronym{gbscw}{gbscw}{Weighted G Brier score}

\newacronym{bl}{bl}{Bernoulli likelihood}
\newacronym{bll}{bll}{Bernoulli log likelihood}
\newacronym{auc}{auc}{area under curve}
\newacronym{km}{km}{Kaplan-Meier}

\newacronym{gan}{gan}{Generative Adversarial Network}

\newacronym{support}{support}{Study to Understand Prognoses Preferences Outcomes and Risks of Treatment}
\newacronym{metabric}{metabric}{Molecular Taxonomy of Breast Cancer International Consortium}
\newacronym{rott}{rott}{Rotterdam Tumor Bank}
\newacronym{gbsg}{gbsg}{German Breast Cancer Study Group}
\newacronym{rott-gbsg}{rott. \& gbsg}{Rotterdam \& GBSG}
\newacronym{flchain}{flchain}{The Assay Of Serum Free Light Chain}
\newacronym{nwtco}{nwtco}{National Wilm’s Tumor Study}
\newacronym{crash2}{crash-2}{Clinical Randomization of an Antifibrinolyticin Significant Hemorrhage 2}
\newacronym{uw-dcal}{uw-dcal}{uniform-weighted d-cal}
\newacronym{ipcw-dcal}{ipcw-dcal}{\gls{ipcw} d-cal}
\newacronym{ipcw-xcal}{ipcw-xcal}{\gls{ipcw} x-cal}
\newacronym{ibs}{ibs}{integrated brier score}
\newacronym{ipcw-ibs}{ipcw-ibs}{\gls{ipcw}-\gls{ibs}}

\newacronym{crps}{crps}{continuous ranked probability score}
\newacronym{s-crps}{s-crps}{Survival-\acrshort{crps}}
\newacronym{ifd}{ifd}{individual failure distribution}

\newacronym{hl}{hl}{Hosmer-Lemeshow}
\newacronym{gb}{gb}{Grønnesby-Borgan}
\newacronym{dn}{dn}{D’Agostino-Nam}

\newacronym{ni}{ni}{Not-Interpolated}
\newacronym{i}{i}{Interpolated}
\newacronym{mimic}{mimic}{Medical Information Mart for Intensive Care}
\newacronym{mnist}{mnist}{Modified National Institute of Standards and Technology database}
\newacronym{tcga}{tcga}{The Cancer Genome Atlas}
\newacronym{mtlr}{mtlr}{Multi-Task Logistic Regression}
\newacronym{aft}{aft}{Accelerated Failure Times}

\newacronym{icu}{icu}{intensive care unit}

\newacronym{mi}{mi}{Mutual Information}
\newacronym{ipm}{ipm}{Integral Probability Metric}

\newacronym{mmd}{mmd}{Maximum Mean Discrepancy}
\newacronym{hsic}{hsic}{Hilbert-Schmidt Independence Criterion}
\newacronym{mle}{mle}{maximum likelihood estimation}
\newacronym{reg}{reg}{regression}
\newacronym{obs}{obs}{observed}
\newacronym{full}{full}{full}
\newacronym{none}{none}{none}
\newacronym{acc}{acc}{acc}
\newacronym{tr}{tr}{tr}
\newacronym{te}{te}{te}
\newacronym{dr}{dr}{\textit{doubly-robust}}
\newacronym{bmi}{bmi}{body mass index}

\renewcommand{\mid}{~\vert~}

\newcommand{\g}{\mid}

\crefname{lemma}{lemma}{lemmas}
\crefname{prop}{proposition}{propositions}

\DeclareRobustCommand{\indicator}[1]{\ensuremath{\mathbbm{1}\left[#1\right]}}

\newcommand{\indep}{\rotatebox[origin=c]{90}{$\models$}}
\newcommand{\nindep}{\rotatebox[origin=c]{90}{$\not\models$}}

\DeclareMathSymbol{\cat}{\mathord}{operators}{"3A}

\usepackage{algorithm}
\usepackage{algorithmic}
\usepackage{centernot}

\usepackage{pgf,tikz}
\usetikzlibrary{positioning}

\newtheorem*{assumption*}{Assumption}

\newtheorem{proposition}{Proposition}
\newtheorem*{proposition*}{Proposition}
\usepackage{bbm}
\usepackage{siunitx}
\usepackage{amssymb}
\usepackage[footnotes,definitionLists,hashEnumerators,smartEllipses,hybrid]{markdown}
\usepackage{graphicx}
\usepackage{color}
\definecolor{MidnightBlue}{rgb}{0.1, 0.1, 0.44}
\usepackage{soul}

\title[Learning Invariant Representations with Missing Data]{Learning Invariant Representations with Missing Data}
\usepackage{times}

\clearauthor{%
 \Name{Mark Goldstein} \footnote{Work done in part during an internship at Apple.}\Email{goldstein@nyu.edu } \\
 \Name{Aahlad Puli}  \Email{aahlad@nyu.edu}\\
 \Name{Rajesh Ranganath}  \Email{rajeshr@cims.nyu.edu}\\
 \addr New York University
 \AND
 \Name{Jörn-Henrik Jacobsen} \Email{jhjacobsen@apple.com}\\
 \Name{Olina Chau} \Email{olina@apple.com}\\
  \Name{Adriel Saporta} \Email{asaporta@apple.com}\\
\Name{Andrew C. Miller} \Email{acmiller@apple.com}\\
 \addr Apple
}

\begin{document}

\maketitle

\begin{abstract}%
Spurious correlations allow flexible models to predict well during training but poorly on related test distributions. Recent work has shown that models that satisfy particular independencies involving correlation-inducing \textit{nuisance} variables have guarantees on their test performance. Enforcing such independencies requires nuisances to be observed during training. However, nuisances, such as demographics or image background labels, are often missing. Enforcing independence on just the observed data does not imply independence on the entire population. Here we derive \acrshort{mmd} estimators used for invariance objectives under missing nuisances. On simulations and clinical data, optimizing through these estimates achieves test performance similar to using estimators that make use of the full data.
\end{abstract}

\begin{keywords}%
  invariant representations, missing data, doubly robust estimator, MMD
\end{keywords}

\section{Introduction}

Spurious correlations allow models that predict well on training data to have worse than chance performance on related distributions at test time \citep{geirhos2020shortcut,puli2021predictive,veitch2021counterfactual,makar2021causally,gulrajani2020search,sagawa2020investigation}. 
For example, diabetes is associated with high \gls{bmi} in the United States. 
However, in India and Taiwan, diabetes also frequently co-occurs with low and average \gls{bmi} \citep{consultation2004appropriate}.
Due to their shifting relationship with the label, nuisance variables (e.g., \gls{bmi}) can cause models to exploit non-causal correlations in training data, leading models to generalize poorly.

\textit{Invariant prediction} methods are designed to improve performance on a range of test distributions when training data exhibits spurious correlations \citep{peters2016causal,arjovsky2019invariant}.  We focus on methods that enforce independencies between the model and nuisance given some assumed causal structure \citep{makar2021causally,veitch2021counterfactual,puli2021predictive}.
These methods require the nuisance to be specified explicitly and observed.
Nuisances must be observed for all samples because independence constraints 
are enforced via metrics such as \gls{mmd}, which require samples from the fully-observed data. However, in large health datasets, nuisances are often missing. For example, not all people who report diabetes status report other correlated conditions (e.g., hypertension, depression) or demographics (e.g., gender).
To improve generalization on a range of test distributions, it is necessary to handle missingness appropriately. 

We propose \gls{mmd} estimators for measuring
nuisance-model dependence under nuisance missingness. First, we show
that enforcing independence on only the nuisance-observed data does not imply independence on the full data distribution,
and vice versa.
Next, we derive three estimators, including one that is 
\textit{doubly-robust}: it is
consistent when either the nuisance or its missingness can be consistently modeled
given covariates
\citep{bang2005doubly}. Using simulations,
a semi-simulation based on textured \acrshort{mnist}
and clinical data from \acrshort{mimic}, we show that the estimators 
perform close to ground-truth estimation with no missingness and that they
improve test accuracy relative to 
computing the original objective only on samples with nuisances observed.

\section{Notation and background}

\paragraph{Notation.}
Let $X$ denote features.
Let $Y$ be a label such as disease status.
Let $Z$ denote the nuisance, e.g.,
another disease correlated with $Y$, demographics, or image backgrounds.
Denote the nuisance missingness indicator as $\Delta$. 
Instead of $(X,Y,Z)$, we observe $(X,Y,\Delta,\tilde{Z}=\Delta Z)$, where $\tilde{Z}=Z$ when $\Delta=1$
and $Z$ is unobserved otherwise. 
We write functions 
as $f_X=f(X)$.
Let $h_X$ denote a model to predict $Y$. 
When conditioning on events involving both $X,Y$
we use $W=(X,Y)$ and $w=(x,y)$ to lighten notation
so that $\E[\Delta|X=x,Y=y] = \E[\Delta|W=w]$ and 
$f(X,Y)=f_{XY}=f_W$.
Let $\mathcal{B}(p)$ denote $\text{Bernoulli}(p)$.

\paragraph{Modeling under spurious correlations.} Nuisance-based prediction arises in training data 
when $Z$ is predictive of $Y$ and associated with $X$, causing models to use information about $Z$ in $X$ to predict $Y$. 
This may not be a problem in all scenarios, but it is when the test distribution is expected to have a different $(Y,Z)$ relationship from the training distribution, as this may imply
$P_{train}(Y \g X) \neq P_{test}(Y|X)$ and a model built on training data may not perform well on test data. Most approaches to this problem start off by 
narrowing down the set of possible test distributions and their relationship to the training distribution.
In this work, we focus on a family
studied in \cite{makar2021causally,veitch2021counterfactual,puli2021predictive}. The distributions are indexed by $D$ and vary only in the factor
$P(Z \g Y)$: 
\begin{align}
    \label{eq:choiceF}
    \mathcal{F} = \{P_D(X,Y,Z)=P(Y)P_D(Z|Y)P(X|Y,Z)\}_D.
\end{align}
When $P_{test}(Z|Y) \neq P_{train}(Z|Y)$, in general $P_{train}(Y \g X) \neq P_{test}(Y|X)$
and a model built on $P_{train}$ can generalize poorly.
For example consider, for $a \in \mathbb{R}$,
\begin{align*}
   Y \sim \mathcal{B}(0.5), \quad \mu_Z=a(2Y-1), \quad Z \sim \mathcal{N}(\mu_Z,1).
\end{align*}
When 
$P_{train}$ uses $a=0.5$ and $P_{test}$ uses $a=-0.9$,
 \cite{puli2021predictive} show an example of an $X|Y,Z$
distribution
where
an ERM model that predicts $Y$ from $X$ with $80\%$ training accuracy
only achieves $40\%$ on the test set, due to the changing relationship between $Y,Z$, even when $Z$ is not an input feature to the model.

When it is possible to anticipate and observe nuisances during training, enforcing certain independence constraints  \citep{makar2021causally,veitch2021counterfactual,puli2021predictive} helps guarantee performance regardless of the nuisance-label relationship.
For example, for this choice of $\mathcal{F}$ and model $h$, 
maximum likelihood estimation for $Y|h_X$ while
enforcing the constraint $h_X \indep Z \g Y$
implies equal performance on all $P_D \in \mathcal{F}$, and 
better than chance performance
(\Cref{appsec:invariantpredictor}).

\paragraph{Measuring Dependence.}
To enforce independencies it is necessary to measure dependence and to minimize this measure.
One way to measure dependence is to measure distance between a joint distribution and the product of its marginals. The general \glspl{ipm} class defines a metric on distributions. A special case, the kernel-based \gls{mmd} \citep{gretton2012kernel}, has a closed form.
Let $X_1 \sim P, X_2 \sim Q$. Let $X_j^\prime$ be an independent sample identically distributed as $X_j$.
For kernel $k$,
\begin{align}
\label{eq:mmd}
\gls{mmd}_k(P,Q)=
    \E[k(X_1,X_1^\prime)]
    +
    \E[k(X_2,X_2^\prime)]
     - 2 
    \E[k(X_1,X_2)] \,.
\end{align}
The \gls{mmd} is $0$ if and only if $P=_d Q$ under certain mild conditions
on $P,Q$ and the kernel $k$ \citep{gretton2012kernel}.
Computing the \gls{mmd} on a joint distribution $P$ of two variables and the product of its marginals $Q$ is a measure dependence, which also coincidences with the \gls{hsic}
\citep{gretton2005measuring,szabo2017characteristic}.

\begin{figure}[t!]
    \centering
        \includegraphics[width=40mm]{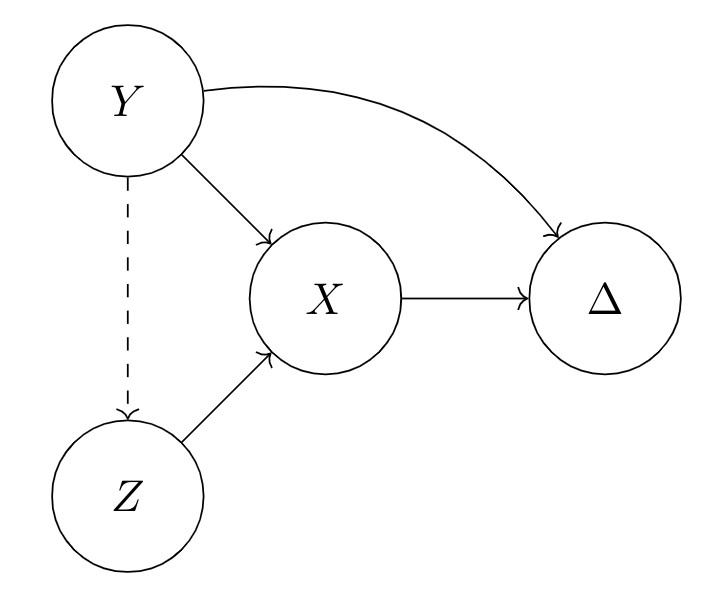}
    \caption{Generative process we consider in this work. The $Y\rightarrow Z$ edge is dashed to emphasize that the $Z|Y$ may change at test time. $\Delta$ determines missingness of $Z$
    and satisfies $Z \indep \Delta | (X,Y)$.     \label{fig:graphs} }
\end{figure}

\paragraph{Assumptions and scope.}
The estimators in this work
require ignorability: $Z \indep \Delta | W$ 
\citep{hernan2021casual}.
This holds, e.g., in graphs such as \Cref{fig:graphs}.
We also require positivity $0 < \epsilon \leq P(\Delta=1|W)$
to observe $Z$ appropriately.
While we focus on the graph in \Cref{fig:graphs} with binary $Z$,
the presented method can extend to 
continuous $Z$ (\Cref{appsec:mmdcontinuous}), 
to other
data generating processes, and to other invariance objectives.
The method applies when the data distributions
satisfy
ignorability, positivity, and any assumptions made by the underlying invariance method.

\paragraph{Estimation under missingness.}
The problem that this work tackles is enforcing conditional independencies such as $h_X \indep Z | Y$ to improve generalization in families like $\mathcal{F}$
in \Cref{eq:choiceF}
even when $Z$ is subject to missingness.
Methodology from causal inference solves a related problem:
estimating $\E[Z]$ when $Z$ is subject to missingness.
In this work, we extend this methodology
from estimates of $\E[Z]$ to estimates of objectives that enforce
independencies involving $Z$.
Here we review estimation of $\E[Z]$ under missingness.
The following assumes ignorability and positivity.
Two parts of the data-generating distribution can help.
Letting $W=(X,Y)$, define:
\begin{align*}
    G_W &\triangleq \mathbb{E}[\Delta \mid X,Y]
    \quad \quad \text{(missingness process)}\\
    m_W &\triangleq 
   \mathbb{E}[Z \mid X,Y]
   \quad \quad \text{(conditional expectation)}\\
\end{align*}
We review estimators of $\E[Z]$ that use $G_W$
\citep{horvitz1952generalization,binder1983variances,robins1994estimation}
or $m_W$ 
\citep{rubin1976inference,schafer1997analysis} 
in \Cref{appsec:ipestimator}.
The \gls{dr} estimator
\citep{robinsrotnitzky2001comment,bang2005doubly,kang2007demystifying}
combines both by noting
the following equality:
\begin{align}
    \label{eq:drestimator}
    \E[Z] = \E
    \Big[ 
        \frac{\Delta \tilde{Z}}{G_W}
        -
        \frac{\Delta - G_W}{G_W}m_W
    \Big] \,.
\end{align}
Crucially, the right side of  \Cref{eq:drestimator}
does not requires samples of $Z$ when $\Delta=0$.
When $G_W$ or $m_W$ are replaced with estimates $\hat{G}_W,\hat{m}_W$,    
the equality still holds provided that for all $w$,
\textit{either} $\hat{G}_w=G_w$
\textit{or} $\hat{m}_w=m_w$
(\Cref{appsec:dr}).
Moreover, Monte Carlo estimates of the right side of \Cref{eq:drestimator} are consistent for $\E[Z]$ when \textit{either} $\hat{G}_W$ consistently estimates $G_W$ or $\hat{m}_W$ consistently estimates $m_W$ 
\citep{robinsrotnitzky2001comment}. This is useful 
because in practice neither of $G_W$ nor $m_W$ are known and both must be estimated.
\section{Invariant representations with missing data }
For the graph in \Cref{fig:graphs} and binary $Z$, \cite{veitch2021counterfactual} enforce $h_X \indep Z | Y$ for the predictive model $h$ by
maximizing likelihood
while minimizing the \gls{mmd}:
\begin{align}
\label{eq:objective} 
\max_h \,
 \quad \log p(y|h_X) - \lambda \cdot \sum_{y \in \{0,1\}} \gls{mmd}_k\left(p(h_X|Z=1,Y=y) \,,\, p(h_X|Z=0,Y=y)\right) \, ,
\end{align}
for $\lambda \geq 0$.
The first term is the usual maximum-likelihood objective for predicting $y$ with model $h_X$.
The second term, because $Z$ is binary, enforces $h_X \indep Z | Y$ when minimized. First, we motivate the independence constraint for the choice of assumed family $\mathcal{F}$ in
\Cref{eq:choiceF}.
We then demonstrate what can go wrong when enforcing this \gls{mmd} penalty only on samples where $Z$ is observed. We then derive estimators of the full-data \gls{mmd} under missingness.

\subsection{Conditional Independence implies
equal performance on the anti-causal family}

There are at least two distinct usages of the word \textit{invariance} in the literature. One refers to independence (e.g. of a model to a nuisance or environment variable). The other refers to \textit{invariant risk}, i.e., the risk is the same for all test distributions in some family
\citep{arjovsky2019invariant,krueger2021out}. In some  families and for some independence constraints, these can
coincide.
\begin{proposition}
    \label{prop:equal} 
    Suppose model $h_X$ satisfies
    $h_X \indep Z | Y$ on any $P_D \in \mathcal{F}$.
    Then for all $P_{D^\prime} \in \mathcal{F}$,
    $\E_{P_D}[\log p_D(Y|h_X)]=\E_{P_{D^\prime}}[\log p_D(Y|h_X)]$.
\end{proposition}
This has been shown
in \cite{veitch2021counterfactual,puli2021predictive}
but we provide a self-contained proof in \Cref{appsec:invariantpredictor}.
This result means that estimates of held-out performance from the training data (one member of $\mathcal{F}$)
will represent test performance on other member of $\mathcal{F}$ under
$h_X \indep Z | Y$.

\subsection{Failures of restricting to observed data \label{sec:observedonlybad}} 

Under missingness, we observe $(X,Y,\Delta,\tilde{Z})$
instead of $(X,Y,Z)$, where $\Delta=1$ means $\tilde{Z}=Z$ and $Z$ is unobserved when $\Delta=0$. When $Z$ is subject to missingness, we cannot directly compute
empirical estimates of the \gls{mmd}. 
What happens when we compute the \gls{mmd} only on samples
with $Z$ observed?
Let us refer to this as the \textit{observed-only} \gls{mmd}.
Restricting computation to data with non-missing $Z$
enforces 
  $h_X \indep Z | Y=y,\Delta=1$ instead of 
   $h_X \indep Z | Y=y$.
We show that these
conditions
do not imply each other in general.
\begin{proposition}
\label{prop:prop1}
There exist distributions on $(X,Y,\Delta,Z)$ such that
\begin{align*}
\exists h_X^\star \quad \text{s.t.} \quad 
    h_X^\star \indep Z | Y=y, 
    \quad 
    \text{but}
    \quad 
    h_X^\star \nindep Z | Y=y,\Delta=1
\end{align*}
and  there exist distributions on $(X,Y,\Delta,Z)$ such that
\begin{align*}
\exists h_X^\star \quad \text{s.t.} \quad 
 h_X^\star \indep Z | Y=y,\Delta=1
 \quad 
 \text{but}
 \quad 
 h_X^\star \nindep Z | Y=y
\end{align*}
\end{proposition}
The proof is in \Cref{appsec:oobad}.
This existence implies:
    \begin{enumerate}
        \item Optimizing
the observed-only \gls{mmd} can discard a
solution to the full-data \gls{mmd}
\item Using the observed-only \gls{mmd} may lead one to believe a model is invariant when it is not.
    \end{enumerate}
To keep generalization guarantees one must enforce independence on the \textit{full data distribution}.

\subsection{\Gls{mmd} estimation under missingness \label{sec:main}}
We present estimators of the full-data
\textit{unconditional} \gls{mmd} under missing $Z$ , which enforces
$h_X \indep Z$ (unconditional on $Y$). Everything that follows
can be conditioned on $Y=y$ to enforce $h_X \indep Z | Y=y$ simply by restricting samples used to estimate the expectations to those with $Y=y$. For a kernel $k$ let $k_{X \xp} \triangleq k(h_X,h_{\xp})$.
The \gls{mmd} can be written as:
\begin{align}
    \label{eq:invariancemmd}
    \gls{mmd}_k \big( p(h_X|Z=1), p(h_X|Z=0)\big)
    =
    \E_{\substack{X|Z=1\\
    \xp|\zp=1
    }}
    k_{X \xp}
    +
    \E_{\substack{X|Z=0\\
    \xp|\zp=0
    }}
      k_{X \xp}
    -
    2
    \E_{\substack{X|Z=1\\
    \xp|\zp=0
    }}
      k_{X \xp}
    \,.
\end{align}
Estimation is challenging due to 
missingness in the conditioning set.
For $b,b^\prime \in \{0,1\}$, let $N(b,b^\prime) \triangleq P(Z=b)P(\zp=b^\prime)$
and let $Z_1 \triangleq Z$ and $Z_0 \triangleq 1-Z$. For any of the expectations, the dependence on $Z$ can be re-written with indicators in a joint expectation: 
\begin{align}
\label{eq:fullmmd}
    \E_{\substack{X|Z=b\\ \xp | \zp = b^\prime}}
    k_{X \xp}
    =
     \frac{1}{N(b,b^\prime)}
    \E \Big[ 
            k_{X \xp} \cdot Z_b \cdot Z^\prime_{b^\prime}
    \Big] \,,
\end{align}
Under no missingness, each expectation could be estimated with Monte Carlo.
We now develop three estimators\footnote{By \textit{estimator}, we really mean
that we provide an alternate form of the expectation. The actual 
estimators we propose are empirical Monte Carlo estimates of the presented expectations, with $G$ and $m$ replaced with models.} for these expectations.
We propose simple $G_W$-based and $m_W$-based estimators in
\Cref{eq:reweightingmmd,eq:regressionmmd}
and then a doubly-robust estimator that combines them 
in \Cref{eq:drmmd}.
\begin{proposition}
     \label{prop:ipcw}
    ($G_W$-based re-weighted estimator) Assume positivity, ignorability, and, 
  $\forall w$, $G_w=\E[\Delta|W=w]$.
   Then,
   \begin{align}
   \label{eq:reweightingmmd}
   \E_{\substack{X|Z=b\\
    \xp|\zp=b^\prime
    }}
    \Big[ 
    k_{X \xp}
    \Big]
    &=
\frac{1}{N(b,b^\prime)}
    \E
    \Big[ 
    \frac{\Delta \Delta^\prime  \tilde{Z}_b
                    \tilde{\zp}_{b^\prime}
        }{G_{W \wp}}
    k_{X \xp}
    \Big].
\end{align}
 \end{proposition}
 \begin{proposition}
     \label{prop:reg} 
    ($m_W$-based regression estimator)
     Assume ignorability, and, 
  $\forall w$, $m_w = \E[Z |W=w]$. Then,
\begin{align}
\label{eq:regressionmmd}
\E_{\substack{X|Z=b\\
    \xp|\zp=b^\prime
    }}
    \Big[ 
    k_{X \xp}
    \Big]
    &=
\frac{1}{N(b,b^\prime)}
    \E
    \Big[ 
    m_{Wb} \cdot m_{\wp b^\prime}
    \cdot 
    k_{X\xp}
    \Big].
\end{align}
 \end{proposition}
Let  $m_{W1} \triangleq m_W,\phantom{.} m_{W0} \triangleq 1-m_W$, and $G_{W \wp} \triangleq G_WG_{\wp}$.
\begin{proposition}
    \label{prop:dr} 
(\gls{dr} estimator). Assume positivity, ignorability, and $\forall w$, either
$G_w=\E[\Delta|W=w]$ or $m_w = \E[Z |W=w]$. Then,
\begin{align}
    \label{eq:drmmd}
\E_{\substack{X|Z=b\\
    \xp|\zp=b^\prime
    }}
    \Big[ 
    k_{X \xp}
    \Big]
    &=
    \frac{1}{N(b,b^\prime)
}
    \E
    \Big[ 
    \Big( 
    \frac{
        \Delta
        \Delta^\prime 
        \tilde{Z}_{b}
        \tilde{\zp}_{b^\prime}
        } 
        {G_{W \wp}}
        -
       \frac{\Delta \Delta^\prime - G_{W \wp}}{G_{W \wp}}
       \cdot 
     m_{W b}
     \cdot 
     m_{\wp b^\prime}
    \Big)  k_{X \xp}
    \Big] \,.
\end{align}
\end{proposition}
The proof is
in \Cref{appsec:mmdestimators}. 
We can use any 
of \Cref{eq:reweightingmmd,eq:regressionmmd,eq:drmmd}
to estimate the terms in 
\cref{eq:invariancemmd}.
Each of 
\Cref{eq:reweightingmmd,eq:regressionmmd,eq:drmmd} 
is a ratio of two expectations: the normalization
constant $N(b,b^\prime)$ depends on $\E[Z]$
and must itself be estimated
under missingness
(e.g., with \Cref{eq:drestimator}).
The ratio of consistent estimates 
of these quantities 
is consistent by Weak Law of Large Numbers and Slutsky's theorem.
We discuss estimation in practice, trade-offs among the three estimators, and variance
in \Cref{appsec:estimationinpractice}.
We review recent related work in \Cref{sec:related}.

\section{Experiments} 
We compare accuracy and \gls{mmd} minimization using different estimators:
\underline{\acrshort{none}} (\acrshort{mle} only, no \gls{mmd}), 
\underline{\acrshort{full}} (\acrshort{mle} and \gls{mmd} using data with $Z$ fully-observed, which is
what could be used as an objective under no missingness), 
\underline{\acrshort{obs}}  (\acrshort{mle} and observed-only \gls{mmd}), 
\underline{\gls{dr}} (\acrshort{mle} and \gls{dr} estimator, called \underline{\gls{dr}+} when using true $G_W$),
\underline{\acrshort{ip}} (\acrshort{mle} and re-weighted estimator, called \underline{\acrshort{ip}+} when using
true $G_W$), and
\underline{\acrshort{reg}} (\acrshort{mle} and regression estimator).

We first compare these algorithms in a simulation study. 
We then use 
textured \acrshort{mnist} to show the utility of the proposed estimators
on high-dimensional data.  In quantitative tables, we show mean $\pm$ standard deviation over three seeds.
We then predict hospital length of stay
in the \acrshort{mimic} dataset, and compare performance when demographic nuisances are subject to missingness.
For the $Y|X$ predictive loss, we use negative Bernoulli log likelihood with logit equal to model output $h_X$.

\paragraph{Comparing \glspl{mmd}.} In all tables, the training set \gls{mmd} to evaluate each method is computed using the ground-truth full-data \gls{mmd} estimation method (see \cref{eq:fullmmd}) to show the actual value of \gls{mmd} achieved, regardless of optimization method.
This is also what the 
\acrshort{full} method directly optimizes. True $Z$'s are available in both simulated and real data as missingness is simulated. However, each model trains and validates the $\log p(Y|X) + \gls{mmd}$ loss 
using its own estimation method for \gls{mmd}.

\subsection{Experiment 1: Simulation.} We set up strong $(Y,Z)$ correlation.
With $\overline{Y}=1-Y$, the training and validation sets are drawn:
\begin{align}
    \label{eq:dgp1}
    \begin{split}
  Y \sim \mathcal{B}(0.5), \quad 
  Z \sim \mathcal{B}(.9Y + .1\overline{Y}), \quad 
  X \sim [\mathcal{N}(Y-Z,\sigma_X^2), \mathcal{N}(Y+Z,\sigma_X^2)] \,.
\end{split}
\end{align}
The test set has the opposite relationship
$Z \sim \mathcal{B}(.1Y + .9\overline{Y})$. 
Here $h_X^\star = (X_1 + X_2)/2$  predicts $Y$ with smallest MSE among representations satisfying independence.
We construct $\Delta$
to show the failure of computing \gls{mmd} on the observed-only subset.
For this, we use $\hat{Z} \triangleq -(X_1-X_2)/2$, which is correlated with $Z$. We draw $\Delta$ conditional on 
$h_X^\star$ and $\hat{Z}$ (both are functions of $X$):
\begin{align*}
    Q=\indicator{h_X^\star >0.6} \cdot \indicator{\hat{Z}<0.6}, \quad \Delta \sim \mathcal{B}(Q + 0.2\overline{Q})\, .
\end{align*}
This example construction leads to $h_X^\star \indep Z | Y$
but $h_X^\star \nindep Z | Y,\Delta=1$.
For $h, G_W$ and $m_W$ we use small feed-forward neural networks. See the repository\footnote{\href{https://github.com/marikgoldstein/missing-mmd}{https://github.com/marikgoldstein/missing-mmd}} for more details.

\paragraph{Results.}  In \Cref{tab:simulation}, the \gls{dr} estimators
achieve indistinguishable performance
to the full-data \gls{mmd}, both in \gls{mmd} and accuracy, and better than \acrshort{none} and \acrshort{obs}. 
We include more results in \Cref{appsec:fullexperiments}.
\begin{table}
\centering
\caption{\label{tab:simulation}  Simulation. $\lambda=1$. 
\Acrshort{none} has highest \gls{mmd} and lowest test accuracy. \Acrshort{obs} improves over this. The \acrshort{dr} and \acrshort{reg} methods are able to bring the \acrshort{mmd} close to $0.0$ and attain best test accuracy.}

\begin{tabular}{llllllllll}
\toprule 
           &   \acrshort{none}        &   \acrshort{obs}          &  \acrshort{full}     & \gls{dr}    & \gls{dr}+
           & \acrshort{reg} 
           \\
\midrule 
\acrshort{tr} \gls{mmd}   &   $0.21 \pm 0.04$        &   $0.05 \pm 0.04$       & $0.00 \pm 0.01$    & $0.00 \pm 0.01$     & $0.00 \pm 0.01$  
&$0.01 \pm 0.00$ 
\\
\acrshort{tr} \acrshort{acc}   &   $0.89 \pm 0.00$       &   $0.87 \pm 0.00$    &   $0.86 \pm 0.01$    &  $0.85 \pm 0.01$       & $0.84 \pm 0.02$  &$0.86 \pm 0.00$ \\
\acrshort{te} \acrshort{acc}    &   $0.67 \pm 0.02$          &   $0.77 \pm 0.02$      &  $0.80 \pm 0.01$     & $0.81 \pm 0.02$      & $0.81 \pm 0.01$ &$0.79 \pm 0.02$ \\
\bottomrule
\end{tabular}
\end{table}
\subsection{Experiment 2: Textured \acrshort{mnist}.}
\begin{figure}
    \centering
    \begin{minipage}[c]{0.60\textwidth}
    \subfigure{
        \includegraphics[width=20mm]{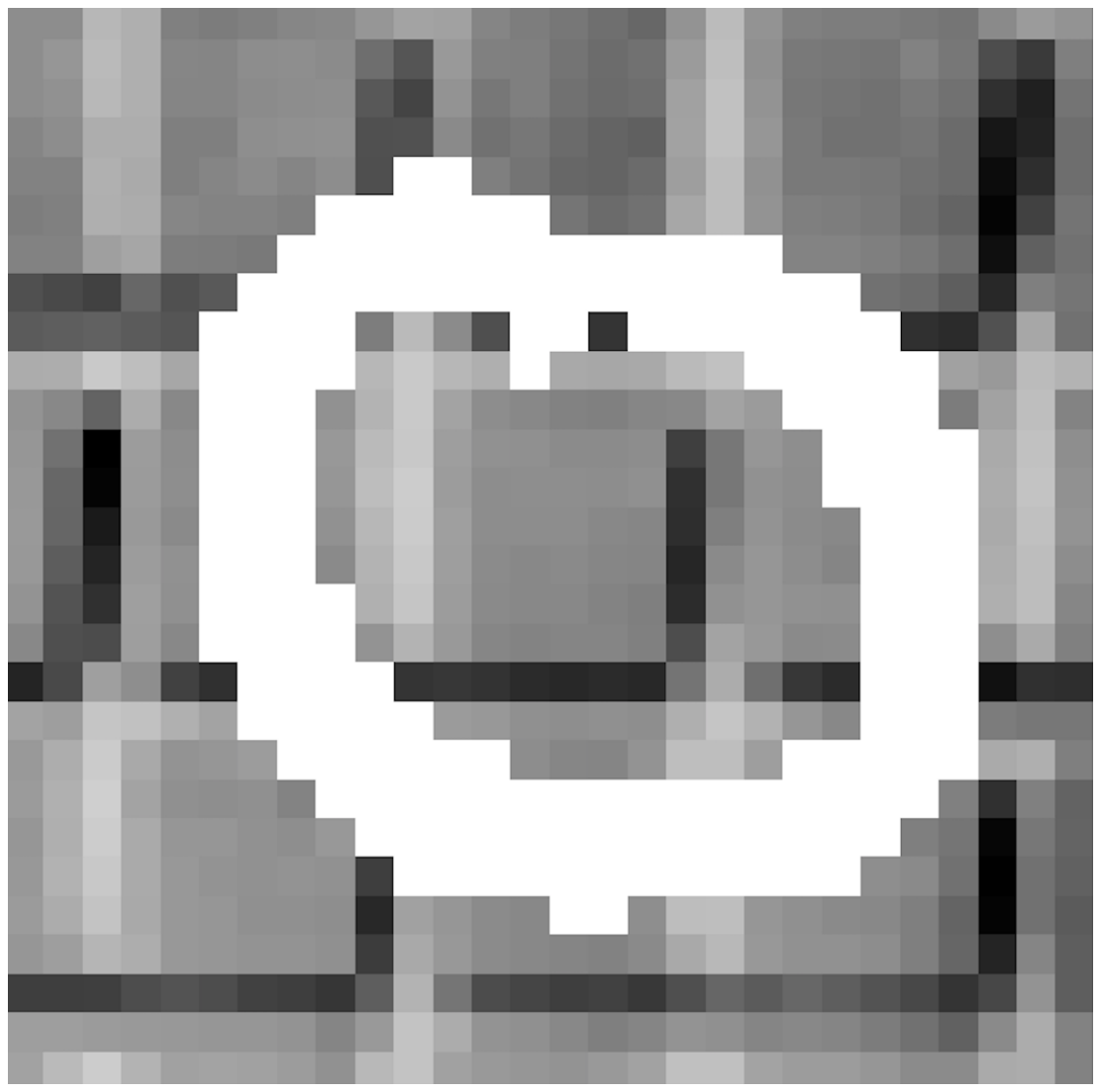}}
    \subfigure{
        \includegraphics[width=20mm]{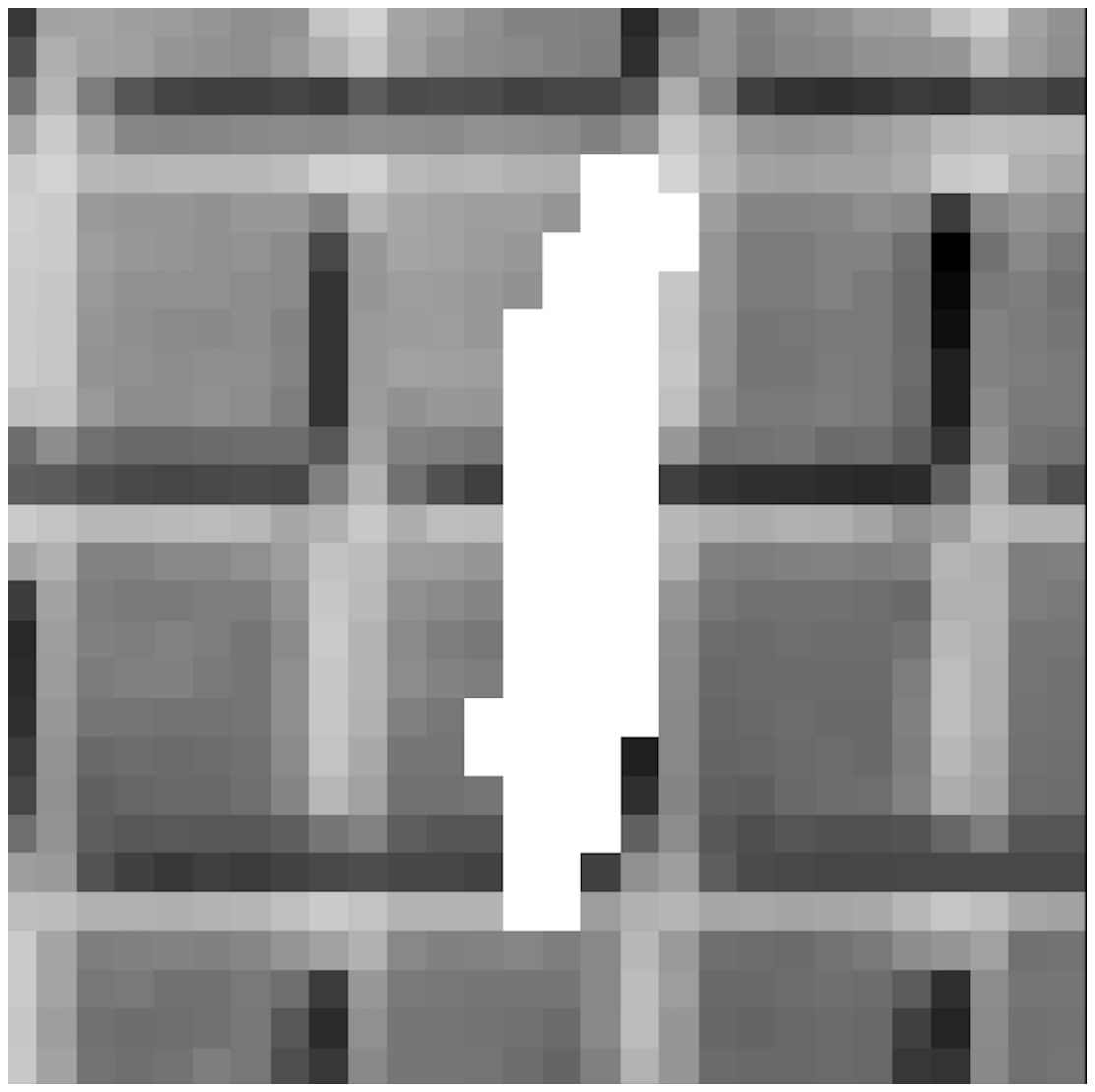}}
    \subfigure{
        \includegraphics[width=20mm]{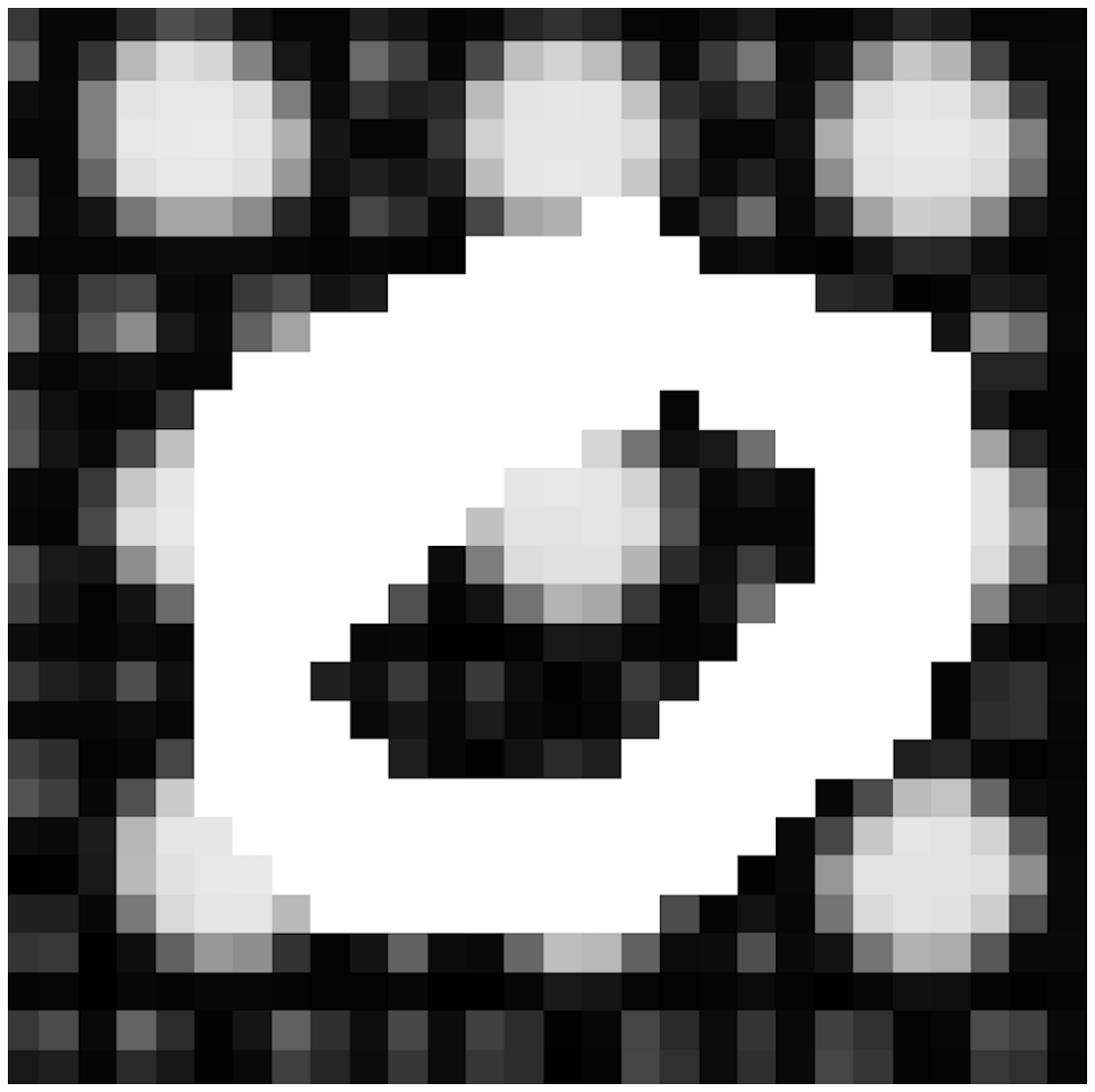}}
    \subfigure{
        \includegraphics[width=20mm]{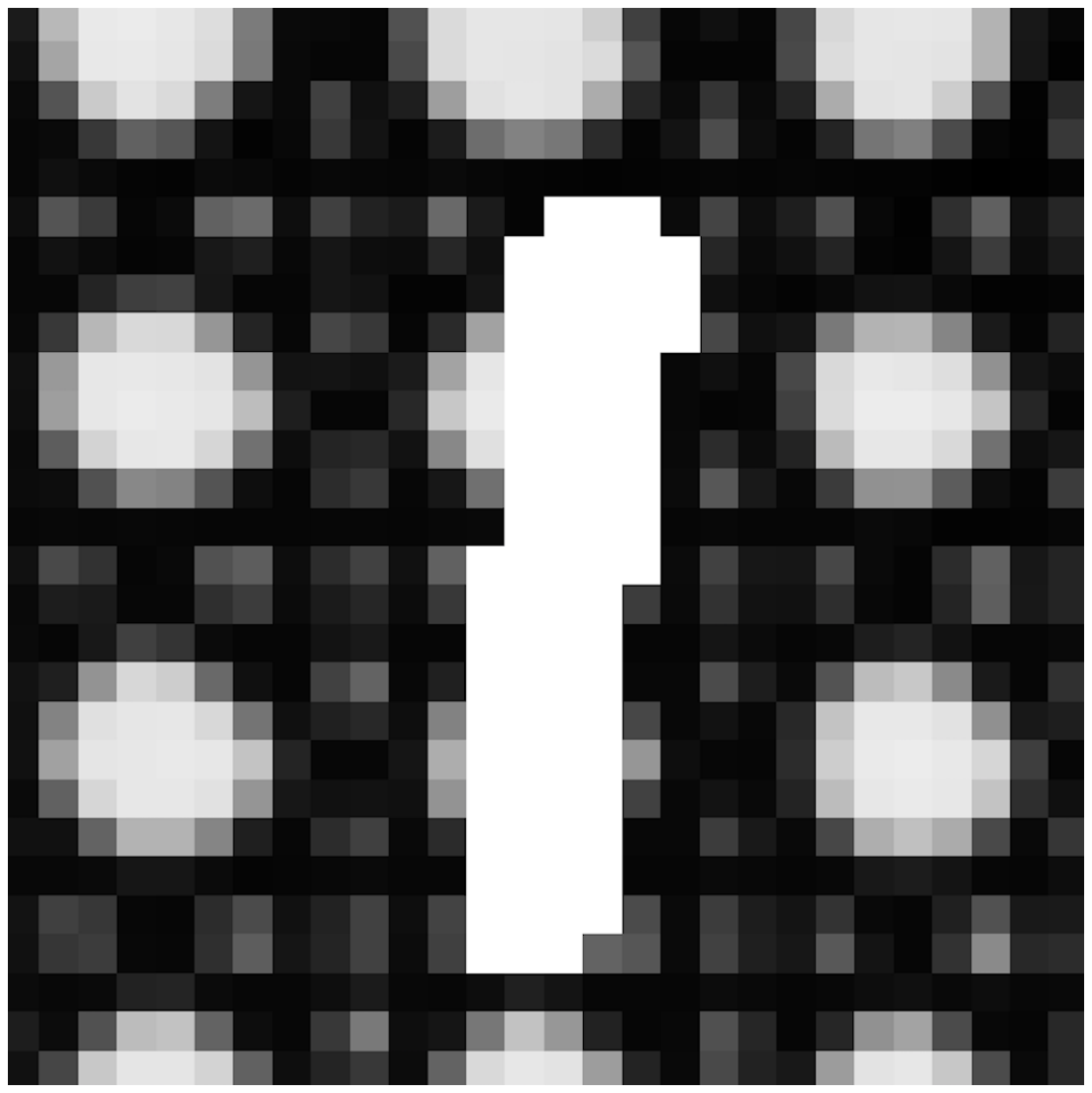}}
    \end{minipage}\hfill
    \begin{minipage}[c]{0.40\textwidth}    
    \caption{Textured \acrshort{mnist} with digits 0,1 on 
    two textures
    from the Brodatz dataset.}
   \label{fig:tmnist}
   \end{minipage}
\end{figure}

Following \cite{pylearn2}\footnote{We adapt \href{https://github.com/lisa-lab/pylearn2/blob/master/pylearn2/scripts/datasets/make_mnistplus.py}{this repository (linked)} to construct textured \acrshort{mnist} and will make our code available.}, 
we correlate \acrshort{mnist} digits $0$ and $1$ with two textures from the Brodatz dataset (\Cref{fig:tmnist}). This is an example of invariance to image backgrounds when not all background labels are available. We follow a similar setup to colored \acrshort{mnist}  \citep{arjovsky2019invariant}: because $Y|X$ is essentially deterministic, even strong spurious correlations may be ignored by a model on \acrshort{mnist}. To push $Y|X$ closer to what may be expected in noisier real data, we flip the label with 25\% chance. Letting $X$ only predict $Y$ with $75\%$ chance means the model will use texture instead. The missingness is 
based on  the average pixel intensity of $X$ and its class. Let $\mu_X$ be the mean pixel value of a 28x28 \acrshort{mnist} image. We set
\begin{align*}
    Q=\indicator{Y=1} \cdot \indicator{\mu_X < 0.3}, \quad \Delta \sim \mathcal{B}(Q + .2\overline{Q}).
\end{align*}
The choice of $Q$ is correlated with $Z$ through whether the image is light or dark grey. Similar to \cref{prop:prop1} and experiment 1, this means subsetting on $\Delta=1$ does not imply independence on the full population and may throw away solutions that do.
For $h, G_W$ and $m_W$ we use 
small convolutional networks. We include more details in the repository.

\paragraph{Results.}  
In \Cref{tab:mnist1}, 
\acrshort{none} and 
\acrshort{obs} perform poorly on test.
In contrast, the \gls{dr} estimators --- including the one with a learned $G_W,m_W$ --- achieve close to \acrshort{full}'s performance. 
\begin{table}
\centering
\caption{\label{tab:mnist1} \acrshort{mnist}  $\lambda=1$.
\gls{dr} and \acrshort{reg} estimators achieve close to full performance as measured by full \gls{mmd}$=0$ and high test accuracy.
\Acrshort{none} and \acrshort{obs} perform poorly on test.
\Acrshort{obs} is notably high variance.}
\begin{tabular}{llllllllll}
\toprule 
           &   \acrshort{none}             &   \acrshort{obs}         &  \acrshort{full}      &\gls{dr}     &\gls{dr}+  & \acrshort{reg} \\
\midrule 
\acrshort{tr} \gls{mmd}  &  $2.05 \pm 0.18$   &  $0.02 \pm 0.04$       &$0.00 \pm 0.01$   & $0.00 \pm 0.01$    & $0.00 \pm 0.01$  & $0.00 \pm 0.01$ 
 \\
\acrshort{tr} \acrshort{acc}    &  $0.90 \pm 0.01$       & $0.74 \pm 0.03$      &$0.76 \pm 0.01$ & $0.77 \pm 0.0$    & $0.76 \pm 0.01$   & $0.76 \pm 0.01$  \\
\acrshort{te} \acrshort{acc}     &  $0.13 \pm 0.01$     & $0.63 \pm 0.17$      & $0.74 \pm 0.01$   & $0.72 \pm 0.04$     & $0.73 \pm 0.01$
 & $0.73 \pm 0.01$\\
\bottomrule
\end{tabular}
\end{table}

\subsection{Experiment 3: Predicting length of stay in the \acrshort{icu}}

We predict length of stay in the \gls{icu} in \acrshort{mimic} \citep{johnson2016mimic}\footnote{The \acrshort{mimic} critical-care database is available on \href{https://physionet.org/}{Physionet}
\citep{goldberger2000physiobank}.}
using demographics and first day labs/vitals
among patients that stay at least one day.
The prediction task is whether the stay is more than $2.5$ days.
To demonstrate that spurious correlations 
cause issues at deployment, we choose 
$Z=1$ to indicate
the patient is recorded as white.
While race may be
correlated with health outcomes (e.g., due to unobserved socioeconomic factors \citep{obermeyer2019dissecting}), it may not always be appropriate for a model to use this information \citep{chen2018my}. The test set represents a
new population with different outcome-demographic structure:
we split the data so that the training/validation set has mostly samples with 
$Y \neq Z$ while the test set has mostly samples with $Y=Z$. We set non-male patients to have $Z$ observed with probability $0.2$. We include more details in the repository.
\begin{table}
\centering
\caption{\label{tab:mimic1} \acrshort{mimic}  $\lambda=1$. \acrshort{reg} estimator matches \acrshort{full}'s performance and improves upon \acrshort{obs} while \gls{dr} does not, due to high objective variance during training (not shown in table).}
\begin{tabular}{lllllllll}
\toprule 
&   \acrshort{none}             &   \acrshort{obs}         &  \acrshort{full}      &\gls{dr}     &\acrshort{reg}  \\
\midrule
\acrshort{tr} \gls{mmd}  & $0.017 \pm 0.02$  &  $0.002 \pm 0.01$       & $0.00 \pm 0.00$    & $0.009 \pm 0.01$    & $0.00 \pm 0.00$  \\
\acrshort{tr} \acrshort{acc}   & $0.71 \pm 0.02$   &  $0.68 \pm 0.01$       &  $0.70 \pm 0.01$    & $0.70 \pm 0.01$    & $0.71 \pm 0.00$  \\
\acrshort{te} \acrshort{acc}     & $0.64 \pm 0.00$  &   $0.64 \pm 0.00$       & $0.66 \pm 0.00$   & $0.62 \pm 0.00$   & $0.66 \pm 0.01$ \\
\bottomrule
\end{tabular}
\end{table}
\paragraph{Results.}  
In this real data setting with strong $(Y,Z)$ correlation, the full-data \gls{mmd} estimator reported in the table for all methods may have high variance. We focus on the attained accuracies. The \acrshort{reg} estimator matches the ground-truth \acrshort{full} estimator and performs better than \acrshort{obs}
and \acrshort{dr}. This is not unexpected, since it is possible for the \acrshort{reg} estimator to be lower variance than \gls{dr} when the true $G_W$ is small or $G_W$ is not modeled well \citep{davidian2005double}, especially under strong $(Y,Z)$ correlation (\Cref{appsec:estimationinpractice}).

\section{Related work} 
\label{sec:related}

We focus on recent work in fairness and invariant prediction on 
missing group/environment labels. Motivated by fairness, \cite{wang2020robust} study a related problem of optimizing invariance-inducing objectives when the protected group label (analogous to our nuisance variables) is noisy.  Given bounds on the level of label noise, this work proposes optimizing an objective based on the \textit{distributionally robust optimization} framework \citep{namkoong2016stochastic}.  Additionally, if given a small amount of true labels the authors suggest fitting a model to de-noise the noisy group labels and re-weight examples in the objective, which is similar in spirit to our work.  In our approach, however, we exploit structural assumptions about the missingness  process to build a doubly-robust estimator of the \gls{mmd} penalty used during optimization.

\cite{lahoti2020fairness} optimize worst-case-over-groups performance without known group labels. They rely on the assumption that groups are \textit{computationally-identifiable} (i.e. that there exists some function on the data that labels their protected group membership) \citep{hebert2018multicalibration} and use a model to identify groups on which performance is worst. They pose an adversarial optimization 
between the group-labeling model --- which searches for groups with poor performance --- and the primary predictive model. Inspired by this work, \cite{creager2021environment} find worst-case group assignments based on an \gls{erm} model that maximizes invariance penalties and \cite{ahmed2020systematic} illustrate that this objective performs well on a wide range of benchmarks.
Relatedly, \cite{liu2021just} run usual \gls{erm} training and then a second iteration of \gls{erm} that upweights the loss for datapoints on which the model performs badly. This identifies groups with bad model performance without explicit group labels. In both of these works, the groups could be seen either as a nuisance variable or as a confounder that correlates the label and some nuisance variable. 
However, in our setting (and in that of \cite{makar2021causally,veitch2021counterfactual,puli2021predictive}), in exchange for being willing to make assumptions on the test distribution family, we do not need to observe
samples with poor model performance at training time (and may not see any) to prevent
sudden decreases in performance on held-out data at test time.

\section{Conclusion}
We present estimators for the \gls{mmd} that extend recent invariant prediction methods 
to missing data.
Unlike prior estimators that only leverage data with nuisances observed, or consider worst-case estimation, the presented estimators of the full data objective
are consistent when either auxiliary model can be learned. 
As we show in \cref{prop:prop1},
estimation of the full data objective is necessary to preserve the theoretical properties of invariant prediction methods.
In the experiments, the \gls{dr} and \acrshort{reg} estimators are able to match full-data \gls{mmd} performance and improve test accuracy relative to the \acrshort{obs} estimator. In practice, we recommend exploring the two simpler proposed estimators (\acrshort{reg} and \acrshort{ip}) in addition to the \gls{dr} estimator and selecting the model based on the validation metric.

 Moving forward, one limitation is that the full-data estimator --- used as ground-truth \gls{mmd} evaluation for the experiments --- may itself have high variance on small datasets with strong nuisance-label correlation. Variance reduction is an important avenue both for optimizing and evaluating with the \gls{mmd} using smaller batch sizes (in our experiments, batch sizes $1500$ for \acrshort{mnist} and $4000$ for \acrshort{mimic} are large). 
Beyond variance reduction, it is a promising direction to apply the methodology in this work to the mutual information objective in \cite{puli2021predictive}, which sidesteps the choice of kernel and may be better suited for continuous and high dimensional nuisances.

\newpage 

\acks{The authors thank Scotty Fleming, Joe Futoma, Leon Gatys, Sean Jewell, Tayor Killian, and Guillermo Sapiro for feedback and discussions.  This work was in part supported by NIH/NHLBI Award R01HL148248, NSF Award 1922658 NRT-HDR: FUTURE Foundations, Translation, and Responsibility for Data Science,
and NSF Award 1815633 SHF.}

\bibliography{clear2022_cameraready}

\clearpage
\appendix

\section{Invariant predictor \label{appsec:invariantpredictor}} 
Here, we show that
$h_X \indep Z | Y=y$ for all $y$ 
implies invariant risk in $\mathcal{F}$.
The proof uses the following insights
\begin{itemize}
    \item Satisfying independence $h_X \indep Z \g Y$ means $P_{train}(h_X|Y,Z) = P_{train}(h_X|Y)$.
\item The assumptions on $\mathcal{F}$ mean $P_{train}(h_X|Y,Z) = P(h_X|Y,Z)$ in any member of the family. 
\item Combined this means $P_{train}(h_X|Y,Z) = P_D(h_X|Y)$ for any $P_D \in \mathcal{F}$ when the model $P_{train}(Y|h_X)$ satisfies $h_X \indep Z |Y$.
\end{itemize}
\begin{proposition*}
    Suppose model $h_X$ satisfies
    $h_X \indep Z | Y$ on any $P_D \in \mathcal{F}$.
    Then for all $P_{D^\prime} \in \mathcal{F}$,
    $\E_{P_D}[\log p_D(Y|h_X)]=\E_{P_{D^\prime}}[\log p_D(Y|h_X)]$.
\end{proposition*}
\begin{proof}
Consider test set performance $\E_{P_{test}(Y,X)}[ \log P_{train}(Y|h_X)]$. By the assumption on the family, by Bayes, and by satisfying the independence constraint:
    \begin{align*}
    \E_{P_{test}(Y,X)}[ \log P_{train}(Y|h_X)]
    &=
    \E_{P_{test}(Y,X)}
   \Big[ \log \frac{P_{train}(h_X|Y)P(Y)}{P_{train}(h_X)} \Big]\\
    &=
    \E_{P_{test}(Y,X,Z)}
    \Big[ \log \frac{P_{train}(h_X|Y,Z)P(Y)}{\E_{P(Y)} \Big [P_{train}(h_X|Y,Z) \Big]}\Big] \\
&=
    \E_{P_{test}(Y,h_X,Z)}
    \Big[ \log \frac{P_{train}(h_X|Y,Z)P(Y)}{\E_{P(Y)} \Big [P_{train}(h_X|Y,Z) \Big]}\Big] \\
&=
    \E_{P_{test}(Y,h_X,Z)}
    \Big[ \log \frac{P(h_X|Y,Z)P(Y)}{\E_{P(Y)} \Big [P(h_X|Y,Z) \Big]}\Big] \\
&=
    \E_{P_{test}(Y,h_X,Z)}
    \Big[ \log \frac{P(h_X|Y)P(Y)}{\E_{P(Y)} \Big [P(h_X|Y) \Big]}\Big] \\
&=
    \E_{P_{test}(Y,h_X)}
    \Big[ \log \frac{P(h_X|Y)P(Y)}{\E_{P(Y)} \Big [P(h_X|Y) \Big]}\Big] \\
&=
    \E_{P_{test}(h_X|Y)P_{test}(Y)}
    \Big[ \log \frac{P(h_X|Y)P(Y)}{\E_{P(Y)} \Big [P(h_X|Y) \Big]}\Big] \\
&=
    \E_{P(h_X|Y)P(Y)}
    \Big[ \log \frac{P(h_X|Y)P(Y)}{\E_{P(Y)} \Big [P(h_X|Y) \Big]}\Big] \\
&=
    \E_{P(h_X,Y)}
    \Big[ \log \frac{P(h_X|Y)P(Y)}{\E_{P(Y)} \Big [P(h_X|Y) \Big]}\Big] 
    \end{align*}
    The last quantity does not depend on any specific $P_D(Z|Y)$.
    This means that performance of the $P_{train}(Y|h_X)$ model,
    when the independence is satisfied, is the same on all $P_{test}$ in $\mathcal{F}$.
\end{proof}
\newpage 

\section{Estimation in practice \label{appsec:estimationinpractice}}

\subsection{Splitting samples}
For a given batch, we use $1/4$ of the samples for the normalization term and $3/4$ for the main term,
though this number may be changed. Further, the main term of any of the three estimators is defined on a pair of independent samples, i.e. it is a \textit{U-statistic}. There are two ways to estimate such expectations. One option is to further break the samples left for the main term in half
into two batches $S_1$ and $S_2$ and then compute on all pairs $i \in S_1, j \in S_2$. The alternative, which has slightly higher sampler efficiency and is the method we use, is to compute on all pairs of samples and then leave out any diagonal terms $k(X_i,X_i)$ from the average.

\subsection{Trade-offs among the 3 proposed estimators \label{appsec:tradeoffs}}

For large samples, \acrshort{dr} estimates with correct $G_W$ and correct $m_W$ are lower variance than the regression with correct $m_W$, and lower variance than re-weighting with correct $G_W$. Even when $m_W$ is mis-specified but $G_W$ is correct, the \gls{dr} estimator may still be lower variance than the re-weighting estimator with correct $G_W$ alone.
However, the \gls{dr} estimator with correct $m_W$ but mis-specified $G_W$ may be \textit{higher variance} than the regression estimator with correct $m_W$ \citep{davidian2005double}. For this reason, when the missingness model $G_W$ is wrong, the regression estimator may out-perform the \gls{dr} estimator even in large sample sizes.

The variance of the \gls{dr} and re-weighting estimators comes from
two distinct places. One is general to missingness: small observation probabilities $G_W$ in the denominator. The other reason is general Monte Carlo error: we need individual samples of $\tilde{Z}$ in the numerator. This is \textit{especially a problem in the spurious correlation setting}:
$Y$ and $Z$ are possibly strongly correlated. We need to compute the \gls{mmd} conditional on $Y=y$ which involves, for each $Y=y$, expectations using samples where $Z=1$ and where $Z=0$, but we may have very few samples for one of these $Z$ values. This second source of variance \textit{also applies to estimates of the full-data} \gls{mmd} under no missingness (\cref{eq:fullmmd}). We compare the mean and variance of these estimators empirically in \Cref{appsec:var}.

\subsection{Empirical investigation of variance \label{appsec:var}}

\begin{figure}
    \centering
    \subfigure[Means]{
    \label{fig:meanplot}
        \includegraphics[height=46mm]{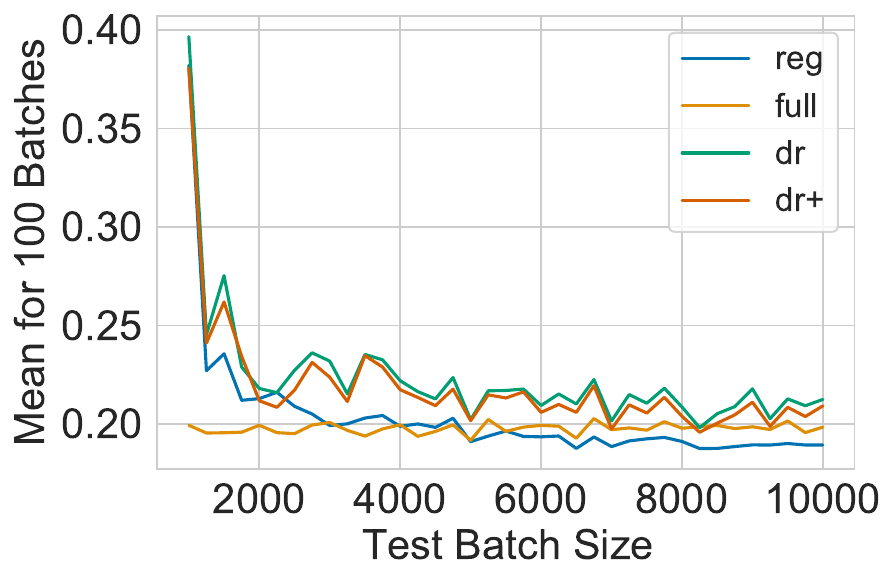}
    }
    \subfigure[Std. Dev.'s]{
    \label{fig:stdplot}
        \includegraphics[height=46mm]{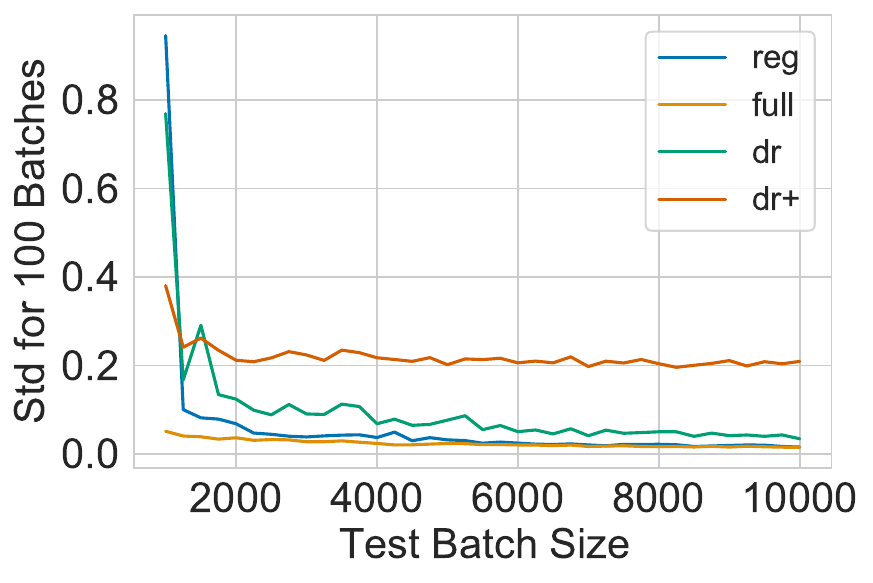}
    }
    \caption{
    \Cref{fig:meanplot}: 
    Mean of 100 MMD estimates at each batch size.
    \Cref{fig:stdplot}: Standard Deviation of 100 MMD estimates at each batch size \label{fig:meanstdplot}
 }
\end{figure}

As discussed, when  $\E[\Delta|X,Y]$ small, or $(Y,Z)$ highly correlated, or both, all estimators will be high variance. We train a model on the experiment 1 simulation using the \acrshort{none} method and then study the mean and variance of \gls{dr}, \gls{dr}+ (to study the effect of using the true $G_W$), \acrshort{reg} (since it yielded better performance on \acrshort{mimic}) and \acrshort{full} (since this method is used to report the \glspl{mmd} in the tables). In this simulation, we are free to generate as many large batches of samples as needed. Keeping the model fixed, for each batch size between $1000$ and $10,000$ incrementing by $250$ we draw $100$ new batches of that size and estimate the \gls{mmd} using each method. For each method, we report the mean (\cref{fig:meanplot}) and standard deviation (\cref{fig:stdplot}) of these estimates.

Notably, we cannot compute an actual ground-truth for the \gls{mmd} of this model, but we could take the mean of the \acrshort{full} estimate (no missingness) at the largest sample size of $10,000$ samples. This is about $0.2$.
We see that the regression estimator stays closer to this number for all sample sizes relative to the \gls{dr} methods. Interesting, for standard deviation, we see that the \gls{dr} estimator is more well-behaved than the \gls{dr}+ estimator that uses the true $G_W$. This has also been observed for learned versus true propensity scores in treatment effect estimation and usually results from models learning less extreme probabilities than the true ones, trading some bias. In this case, there is not a substantial difference in estimated weights or in bias, but there is a large difference in variance. More investigation is required.

The main take-away from both plots is that the regression method seems more stable than \gls{dr} and that $G_W$ may be the part of the \gls{dr} estimator that is not being learned well. On the other hand the \gls{dr} estimator may possibly be safer when it is unknown if it is easier to estimate $G_W$ or $m_W$. We recommend using all $3$ of the proposed estimators and comparing validation objectives.

\section{Full experiments \label{appsec:fullexperiments}}

\begin{table}[h!]
\small 
\centering
\caption{Simulation. $\lambda=1$.}
\label{tab:fullexp1lam1}
\begin{tabular}{lllllllll}
\toprule 
           &   \acrshort{none}        &   \acrshort{obs}          &  \acrshort{full}     & \gls{dr}    & \gls{dr}+  
         &\acrshort{reg} & \acrshort{ip}       & \acrshort{ip}+ \\
\midrule 
\acrshort{tr} \gls{mmd}   &   $0.21 \pm 0.04$&   $0.05 \pm 0.04$& $0.00 \pm 0.01$   & $0.00 \pm 0.01$    & $0.00 \pm 0.01$ &$0.01 \pm 0.00$ &$0.00 \pm 0.00$&$0.00 \pm 0.01$   \\
\acrshort{tr} \acrshort{acc}   &   $0.89 \pm 0.00$ &   $0.87 \pm 0.00$   &   $0.86 \pm 0.01$    &  $0.85 \pm 0.01$     & $0.84 \pm 0.02$ &$0.86 \pm 0.00$&$0.84 \pm 0.01$ & $0.84 \pm 0.01$   \\
\acrshort{te} \acrshort{acc}    &   $0.67 \pm 0.02$ &   $0.77 \pm 0.02$  &  $0.80 \pm 0.01$     & $0.81 \pm 0.02$    & $0.81 \pm 0.01$&$0.79 \pm 0.02$&$0.82 \pm 0.02$&$0.81 \pm 0.00$  \\
\bottomrule
\end{tabular}
\end{table}

\begin{table}[h!]
\small
\centering
\caption{Simulation. $\lambda=5$.}
\label{tab:fullexp1lam5}
\begin{tabular}{lllllllll}
\toprule 
           &   \acrshort{none}        &   \acrshort{obs}          &  \acrshort{full}     & \gls{dr}    & \gls{dr}+  
         &\acrshort{reg} & \acrshort{ip}       & \acrshort{ip}+ \\
\midrule 
\acrshort{tr} \gls{mmd}  &$0.21 \pm 0.04$&$0.03 \pm 0.02$&$0.00 \pm 0.01$&$0.00 \pm 0.00$ &$0.00 \pm 0.0$&$0.00 \pm 0.01$&$0.00 \pm 0.0$ &$0.00 \pm 0.0$\\
\acrshort{tr} \acrshort{acc}   &$0.89 \pm 0.0$ &$0.85 \pm 0.02$&$0.84 \pm 0.01$  &$0.82 \pm 0.01$&$0.78 \pm 0.06$ &$0.84 \pm 0.00$&$0.81 \pm 0.02$&$0.81 \pm 0.03$ 
   \\
\acrshort{te} \acrshort{acc}   &$0.67 \pm 0.02$&$0.78 \pm 0.02$&$0.83 \pm 0.01$&$0.82 \pm 0.02$&$0.77 \pm 0.04$&$0.82 \pm 0.01$&$0.81 \pm 0.02$&$0.80 \pm 0.01$
   \\
\bottomrule
\end{tabular}
\end{table}

\begin{table}[h!]
\small
\centering
\caption{ \acrshort{mnist}  $\lambda=1$.}
\label{tab:fullexp2lam1}
\begin{tabular}{lllllllll}
\toprule 
           &   \acrshort{none}             &   \acrshort{obs}         &  \acrshort{full}      &\gls{dr}     &\gls{dr}+   &\acrshort{reg} & \acrshort{ip}       & \acrshort{ip}+ \\
\midrule 
\acrshort{tr} \gls{mmd}  &  $2.05 \pm 0.18$   &  $0.02 \pm 0.04$       &$0.00 \pm 0.01$   & $0.00 \pm 0.01$  
 &$0.00 \pm 0.01$  &$0.00 \pm 0.01$    &$0.07 \pm 0.12$.   &$0.03 \pm 0.06$\\
\acrshort{tr} \acrshort{acc}    &  $0.90 \pm 0.01$       & $0.74 \pm 0.03$      &$0.76 \pm 0.01$ & $0.77 \pm 0.00$    & $0.76 \pm 0.01$    &$0.76 \pm 0.01$      & $0.67 \pm 0.16$
& $0.68 \pm 0.15$ \\
\acrshort{te} \acrshort{acc}     &  $0.13 \pm 0.01$     & $0.63 \pm 0.17$      & $0.74 \pm 0.01$   & $0.72 \pm 0.04$     & $0.73 \pm 0.01$     &$0.73 \pm 0.01$
&$0.64 \pm 0.14$&$0.61 \pm 0.11$ \\
\bottomrule
\end{tabular}
\end{table}

\begin{table}[h!]
\small
\centering
\caption{ \acrshort{mnist}  $\lambda=5$.}
\label{tab:fullexp2lam5}
\begin{tabular}{lllllllll}
\toprule 
           &   \acrshort{none}            &   \acrshort{obs}         &  \acrshort{full}      & \gls{dr}      &\gls{dr}+          &\acrshort{reg} & \acrshort{ip}     &\acrshort{ip}+ \\
\midrule 
\acrshort{tr} \gls{mmd}   &  $2.05 \pm 0.18$  &   $0.01 \pm 0.02$      & $0.00 \pm 0.00$   & $0.00 \pm 0.00$    & $0.00 \pm 0.01$   &$0.00 \pm 0.01$&$0.01 \pm 0.01$ &$0.01 \pm 0.02$ \\
\acrshort{tr} \acrshort{acc}  &$0.9 \pm 0.01$
&$0.66 \pm 0.15$
&$0.75 \pm 0.01$
&$0.65 \pm 0.14$
&$0.65 \pm 0.13$ 
&$0.75 \pm 0.01$ 
&$0.71 \pm 0.08$
&$0.60 \pm 0.12$ 
\\
\acrshort{te} \acrshort{acc} 
& $0.13 \pm 0.01$
& $0.65 \pm 0.15$
&  $0.75 \pm 0.01$
&$0.73 \pm 0.02$
& $0.70 \pm 0.09$
& $0.75 \pm 0.01$
&  $0.55 \pm 0.3$
&  $0.60 \pm 0.12$
\\
\bottomrule
\end{tabular}
\end{table}

\newpage

\section{Failures of restricting to observed data \label{appsec:oobad}} 

\begin{proposition*}
There exist distributions on $(X,Y,\Delta,Z)$ such that
\begin{align*}
\exists h_X^\star \quad \text{s.t.} \quad 
    h_X^\star \indep Z | Y=y, 
    \quad 
    \text{but}
    \quad 
    h_X^\star \nindep Z | Y=y,\Delta=1
\end{align*}
and  there exist distributions on $(X,Y,\Delta,Z)$ such that
\begin{align*}
\exists h_X^\star \quad \text{s.t.} \quad 
 h_X^\star \indep Z | Y=y,\Delta=1
 \quad 
 \text{but}
 \quad 
 h_X^\star \nindep Z | Y=y
\end{align*}
\end{proposition*}
\paragraph{First direction.}
There exist distributions on $(X,Y,\Delta,Z)$ such that
\begin{align*}
\exists h_X^\star \quad \text{s.t.} \quad 
    h_X^\star \indep Z | Y=y, 
    \quad 
    \text{but}
    \quad 
    h_X^\star \nindep Z | Y=y,\Delta=1
\end{align*}
It suffices to illustrate this even when $Z,Y$ are not correlated.
Consider
\begin{align*}
  Y \sim \mathcal{N}(0,1), \quad 
    Z \sim \mathcal{N}(0,\sigma_Z^2),
    \quad
    \epsilon_X \sim \mathcal{N}(0,\sigma_X^2),
    \quad 
    X &= [Y-Z+\epsilon_X,
    Y+Z]
\end{align*}
For $h_X^\star=(X_1 + X_2)$,
we first show  $h_X^\star \indep Z | Y=y$.
We have 
\begin{align*}
    h_X^\star|Y \sim \mathcal{N}(2Y,\sigma^2_X)
\end{align*}
and in particular $h_X^\star=2Y+\epsilon_X$.
Given $Y=y$, the only randomness in $h_X^\star$
is due $\epsilon_X$. But $\epsilon_X$ is independent of the joint variable $(Z,Y)$ meaning
$\epsilon_X \indep Z | Y=y$ and 
therefore
$h_X^\star \indep Z |Y=y$. 

We now construct $\Delta|(X,Y)$ such that
$h_X^\star \nindep Z | Y=y,\Delta=1$. Let 
\begin{align*}
    \Delta =
\text{OR}\Big( \indicator{X_1+X_2<0}, \indicator{X_2-Y<0}\Big).
\end{align*}
Checking the condition
\begin{align*}
    h_X^\star \nindep Z | Y=y, \Delta=1
\end{align*}
(using definition of $h_X^\star$) is equivalent to checking 
\begin{align*}
    (X_1+X_2) \nindep Z | Y=y,\Delta=1
\end{align*}
(using definition of $\Delta$) is equivalent to checking
\begin{align*}
  (X_1+X_2) \nindep Z | Y=y,
\text{OR}\Big( \indicator{X_1+X_2<0}, \indicator{X_2-Y<0}\Big)=1
\end{align*}
(using defintion of $X_2$) is equivalent to checking
\begin{align*}
    (X_1+X_2) \nindep Z | Y=y,
\text{OR}\Big( \indicator{X_1+X_2<0}, \indicator{Z<0}\Big)=1
\end{align*}
To check that, 
we need to check if the distribution of
$(X_1+X_2)|Y=y,\Delta=1$
changes when conditioning on different events involving
the random variable $Z$. For example, $\indicator{Z<0}$ and $\indicator{Z \geq 0}$:
\begin{enumerate}
    \item $(X_1+X_2) \quad  | \quad Y=y,
\text{OR}\Big(\indicator{X_1+X_2<0},\indicator{Z<0}\Big)=1, \indicator{Z<0}=1$
    \item $(X_1+X_2) \quad   | \quad Y=y,
\text{OR}\Big(\indicator{X_1+X_2<0},\indicator{Z<0}\Big)=1,
\indicator{Z \geq 0}=1$.
\end{enumerate}
We can show these two conditional variables differ in distribution
simply by showing they differ in support.
The first conditional variable can be full support because the event $\indicator{Z<0}$ satisfies one of the OR conditions leaving the other condition $\indicator{X_1+X_2<0}=\indicator{h_X^\star<0}$ free to take either value.
However, the second conditional variable needs $X_1+X_2=h_X^\star<0$ because $\indicator{Z<0}$ is not satisfied (since we condition on $\indicator{Z \geq 0}=1$) but the OR has to be $1$. These different supports imply the distributions differ.
That the variables differ on two non-measure zero sets
is enough to show dependence. Then $(X_1+X_2) \nindep Z | Y=y, \Delta=1$ which means $h_X^\star \nindep Z | Y=y,\Delta=1$.

\paragraph{Second direction.} 
There exist distributions on $(X,Y,\Delta,Z)$ such that
\begin{align*}
\exists h_X^\star \quad \text{s.t.} \quad 
 h_X^\star \indep Z | Y=y,\Delta=1
 \quad 
 \text{but}
 \quad 
 h_X^\star \nindep Z | Y=y
\end{align*}
Let the data be drawn as
\begin{align*}
  Y \sim \mathcal{N}(0,1), \quad 
  Z \sim \mathcal{B}(0.5),\quad
    X &= [Y-Z,
    Y+Z]
\end{align*}
Let $h_X^\star=\indicator{X_1 \geq 0}$. 
We first show $h_X^\star \nindep Z |Y=y$.
We have 
\begin{align*}
h_X^\star
    &= \indicator{X_1 \geq 0}\\
    &= \indicator{Y - Z \geq 0}
\end{align*}
Given $Y=y$, we ask if the random variable $\indicator{y - Z \geq 0}$ is independent of $Z$. To show dependence, we show that the random variable $\indicator{y-Z \geq 0 }$ changes
in distribution when $Z$ takes on its two values:
\begin{enumerate}
    \item  $\indicator{y - Z \geq 0} | Y=y, Z=0$
    \item  $\indicator{y - Z \geq 0} | Y=y, Z=1$
\end{enumerate}
Suppose $y \in (0,1)$. When $Z=0$ we have that $\indicator{y - Z \geq 0}=1$ with probability one.
When $Z=1$, we have  $\indicator{y - Z \geq 0}=0$ with probability one. Therefore
the variables are dependent.

We now let $\Delta = \indicator{X_1 \geq 0} = \indicator{Y - Z \geq 0}$
and show  $h_X^\star \indep Z | Y=y,\Delta=1$. 
Note that $\Delta(X,Y)=h_X^\star$. We ask whether
\begin{align*}
    \indicator{Y - Z \geq 0} \indep Z | Y=y, \indicator{Y - Z \geq 0}
\end{align*}
The conditioning set fully determines 
the variable $\indicator{Y-Z \geq 0}$ meaning it is a constant
and is therefore independent of $Z$. Therefore $h_X^\star \indep Z | Y=y, \Delta=1$ as desired.

\section{IP and outcome estimators}
\label{appsec:ipestimator}

We review estimation of $\E[Z]$ under missingness. 
Two pieces of the data generation process can help,
the missingness process $G_W$
and the conditional expectation $m_W$ of the missing variable:
\begin{align*}
    G_W \triangleq  \mathbb{E}[\Delta \mid X,Y], \quad \quad 
       m_W \triangleq 
   \mathbb{E}[Z \mid X,Y]
\end{align*}
Inverse-weighting estimators 
use $G_W$
\citep{horvitz1952generalization,binder1983variances,robins1994estimation,van2003unified,hernan2021casual}
\begin{align}
\label{eq:ipestimator}
\begin{split}
    \E[Z]
    &=
    \E_X
    \E_{Z|X}
    [Z]\\
    &=
    \E_{X}
    \E_{Z|X}
    \Big[ 
    \frac{\E[\Delta|X]}{\E[\Delta|X]}
    Z
    \Big] \\
    &=\E_X \E_{Z|X}
    \E_{\Delta|X}
    \Big[ 
        \frac{\Delta Z}{\E[\Delta|X]}
    \Big]\\
&=\E_{X Z \Delta}
    \Big[ 
        \frac{\Delta Z}{\E[\Delta|X]}
    \Big]\\
&=\E_{X \Delta Z}
    \Big[ 
        \frac{\Delta Z}{G_W}
    \Big]
\\
&=\E_{X \Delta Z}
    \Big[ 
        \frac{\Delta \tilde{Z}}{G_W}
    \Big]
\end{split}
\end{align}
This means we can estimate $\E[Z]$ provided that (1) ignorability and positivity hold and (2) $G_W$ is known. $G_W$ can be estimated by regressing $\Delta$ on $X$.  Alternatively,
standardization estimators use $m_W$ 
\citep{rubin1976inference,schafer1997analysis,rubin2004multiple,pearl2009causality,little2019statistical,hernan2021casual}:
\begin{align}
    \E[Z]
    =
     \E_X \Big[ \E[Z|X] \Big]
    =
    \E_X \Big[ \E[Z|X,\Delta=1] \Big]
    = \E[m_W]
  \end{align}
The equality between the middle two terms
means that $m_W$ can be estimated by regressing $\tilde{Z}$ on $X$ just on those samples where $\Delta=1$.
The equality follows from the ignorability assumption $Z \indep \Delta | X,Y$.
\section{DR estimator of mean of Z}
\label{appsec:dr}
The inverse weighting and regression estimators can be combined.
\Cref{eq:drestimator} defines the \gls{dr} estimator of $\E[Z]$
by
\begin{align*}
   \E[Z] = \E
    \Big[ 
        \frac{\Delta \tilde{Z}}{G_W}
        -
        \frac{\Delta - G_W}{G_W}m_W
    \Big] 
\end{align*}
Let us re-write this expectation until we see it equals
$\E[Z]$ when $G$ or $m$ are correct.
\begin{align*}
   & \E
\Big[ 
    \frac{\Delta \tilde{Z}}
    {G_W}
    -
    \frac{\Delta - G_W}{G_W}m_W
\Big]\\
&=
   \E
\Big[ 
    \frac{\Delta Z}
    {G_W}
    -
    \frac{\Delta - G_W}{G_W}m_W
\Big]\\
&=
   \E
\Big[ 
    Z
    +
    \frac{\Delta Z}
    {G_W}
    -
    Z
    -
    \frac{\Delta - G_W}{G_W}m_W
\Big]\\
&=
   \E
\Big[ 
    Z
    +
    \frac{\Delta Z}
    {G_W}
    -
    \frac{G_W}{G_W}Z
    -
    \frac{\Delta - G_W}{G_W}m_W
\Big]\\
&=
   \E
\Big[ 
    Z
    +
    \frac{\Delta - G_W}
    {G_W}Z
    -
    \frac{\Delta - G_W}{G_W}m_W
\Big]\\
&=
   \E
\Big[ 
    Z
    +
    \frac{\Delta - G_W}
    {G_W}(Z-m_W)
\Big]\\
&=
   \E
\Big[ 
    Z
\Big]
    +
    \E
    \Big[ 
    \frac{\Delta - G_W}
    {G_W}(Z-m_W)
\Big]
\end{align*}
The first term is what we want, so we just
have to check if the second term is $0$ when
either $G$ or $m$ are correct. If $G$ is correct (regardless of $m$) then:
\begin{align*}
        \E
    \Big[ 
    \frac{\Delta - G_W}
    {G_W}(Z-m_W)
\Big]
&=
    \E
    \Bigg[ 
    \E 
    \Big[ 
    \frac{\Delta - G_W}
    {G_W}(Z-m_W)
    \Big| 
    X,Z
    \Big]
    \Bigg]\\
&=\E
\Bigg[ 
    \frac{\E[\Delta|X,Z] - G_W}
    {G_W}(Z-m_W)
    \Bigg]\\   
 &=\E
\Bigg[ 
    \frac{\E[\Delta|X] - G_W}
    {G_W}(Z-m_W)
    \Bigg]\\      
 &=\E
\Bigg[ 
    \frac{G_W - G_W}
    {G_W}(Z-m_W)
    \Bigg] = 0
\end{align*}
When $m$ is correct (regardless of $G$):
\begin{align*}
            \E
    \Big[ 
    \frac{\Delta - G_W}
    {G_W}(Z-m_W)
\Big]
&=
    \E
    \Bigg[ 
    \E 
    \Big[ 
    \frac{\Delta - G_W}
    {G_W}(Z-m_W)
    \Big| 
    X,\Delta
    \Big]
    \Bigg]\\
&=
    \E
    \Bigg[ 
    \frac{\Delta - G_W}
    {G_W}(\E[Z|X,\Delta]-m_W)
    \Big| 
    X,\Delta
    \Bigg]\\
&=
    \E
    \Bigg[ 
    \frac{\Delta - G_W}
    {G_W}(\E[Z|X,\Delta]-\E[Z|X,\Delta=1])
    \Big| 
    X,\Delta
    \Bigg]\\
&=
    \E
    \Bigg[ 
    \frac{\Delta - G_W}
    {G_W}(\E[Z|X]-\E[Z|X])
    \Big| 
    X,\Delta
    \Bigg] = 0
\end{align*}

\section{Deriving MMD estimators under missingness \label{appsec:mmdestimators}}

\subsection{Deriving the $G_W$-based re-weighted estimator} 
Here we start at the target quantity and derive the estimator. We give the derivation for $Z=1,Z^\prime=1$. The other cases are analogous.
 \begin{align*}
   & \E_{P(X|Z=1)P(\xp|\zp=1)}
    \Big[
    k_{X \xp}
    \Big] \\    
    &=
    \int_{X,\xp}
    k 
    P(X|Z=1)
    P(\xp|\zp=1)
    dX d\xp \\
        &=
            \frac{1}{P(Z=1)} 
    \frac{1}{P(\zp=1)}
        \int_{X,\xp}
    k 
    P(Z=1,X)
    P(\zp=1,\xp)
    dX d\xp \\
    &=
            \frac{1}{P(Z=1)} 
    \frac{1}{P(\zp=1)}
        \int_{X,\xp}
    k 
    P(Z=1|X)
    P(\zp=1|X)
    P(X)P(\xp)
    dX d\xp \\
    &=
            \frac{1}{P(Z=1)} 
    \frac{1}{P(\zp=1)}
        \int_{X,\xp}
    k 
    \E(Z=1|X)
    \E(\zp=1|X)
    P(X)P(\xp)
    dX d\xp \\
    &=
            \frac{1}{P(Z=1)} 
    \frac{1}{P(\zp=1)}
    \E_{\substack{X,Z\\\xp,Z^\prime}}
    \Big[ 
    k
    \cdot 
    Z
    \cdot 
    \zp
    \Big] \\
    &=
            \frac{1}{P(Z=1)} 
    \frac{1}{P(\zp=1)}
    \E_{\substack{X,Z\\\xp,Z^\prime}}
    \Big[ 
    \frac{\E[\Delta|X]\E[\Delta^\prime|\xp]}{\E[\Delta|X]\E[\Delta^\prime|\xp]}
    k
    \cdot 
    Z
    \cdot 
    \zp
    \Big] \\
    &=
    \frac{1}{P(Z=1)} 
    \frac{1}{P(\zp=1)}
    \E_{\substack{X,\Delta,Z\\\xp,\Delta^\prime,Z^\prime}}
    \Big[ 
    \frac{\Delta \Delta^\prime}{G_W G_{\wp}}
    k
    \cdot 
    Z
    \cdot 
    \zp
    \Big]
\end{align*}

\subsection{Deriving the $m_W$-based standardization estimator} 
Here we start at the target quantity and derive the estimator. We give the derivation for $Z=1,Z^\prime=1$. The other cases are analogous.
\begin{align*}
   & \E_{P(X|Z=1)P(\xp|\zp=1)}
    \Big[
    k_{X \xp}
    \Big] \\    
    &=
    \int_{X,\xp}
    k 
    P(X|Z=1)
    P(\xp|\zp=1)
    dX d\xp \\
        &=
            \frac{1}{P(Z=1)} 
    \frac{1}{P(\zp=1)}
        \int_{X,\xp}
    k 
    P(Z=1,X)
    P(\zp=1,\xp)
    dX d\xp \\
    &=
            \frac{1}{P(Z=1)} 
    \frac{1}{P(\zp=1)}
        \int_{X,\xp}
    k 
    P(Z=1|X)
    P(\zp=1|X)
    P(X)P(\xp)
    dX d\xp \\
    &=
            \frac{1}{P(Z=1)} 
    \frac{1}{P(\zp=1)}
        \int_{X,\xp}
    k 
    \E(Z=1|X)
    \E(\zp=1|X)
    P(X)P(\xp)
    dX d\xp \\
    &=
            \frac{1}{P(Z=1)} 
    \frac{1}{P(\zp=1)}
    \E_{X,\xp}
    \Big[ 
     m_W \cdot m_{\wp} \cdot k
    \Big]
    \end{align*}

\subsection{Deriving the DR estimator} 
Here we start at the estimator and derive the target quantity.
We give the derivation for $Z=1,Z^\prime=1$. The other cases are analogous.

\begin{align*}
&
        \frac{1}{P(Z=1)} 
    \frac{1}{P(\zp=1)}
       \frac{1}{N(N-1)}
     \sum_{i \neq j}
    \Big[ 
        \frac{\Delta_{ij} \tilde{Z}_{ij}}{G_{ij}}
        k_{ij} - 
        \frac{\Delta_{ij} - G_{ij}}{G_{ij}}
        m_{ij}k_{ij}
    \Big] \\
&\approx 
        \frac{1}{P(Z=1)} 
    \frac{1}{P(\zp=1)}
\E_{\substack{X,\Delta,Z\\
    \xp,\Delta^\prime,
    \zp}}
    \Big[ 
    \frac{
        \Delta
        \Delta^\prime 
        \tilde{Z} 
        \tilde{Z}^\prime 
        } 
        {G_W G_{\wp}}
        k
        -
        \frac{\Delta \Delta^\prime
        - G_W G_{\wp}}{G_W G_{\wp}}
        m_W m_{\wp} k
    \Big] \\
      &= 
              \frac{1}{P(Z=1)} 
    \frac{1}{P(\zp=1)}
    \E_{\substack{X,\Delta,Z\\
    \xp,\Delta^\prime,
    \zp}}
    \Big[ 
    \frac{
        \Delta
        \Delta^\prime 
        Z 
        \zp 
        } 
        {G_W G_{\wp}}
        k
        -
        \frac{\Delta \Delta^\prime
        - G_W G_{\wp}}{G_W G_{\wp}}
        m_W m_{\wp} k
    \Big] \\
    &=
            \frac{1}{P(Z=1)} 
    \frac{1}{P(\zp=1)}
    \E_{\substack{X,\Delta,Z\\
    \xp,\Delta^\prime,
    \zp}}
    \Big[ 
    Z
       \zp 
        k
        +
   \frac{
        \Delta
        \Delta^\prime 
        } 
        {G_W G_{\wp}}
         Z\zp 
        k
        -
        Z\zp 
        k
        -
        \frac{\Delta \Delta^\prime
        - G_W G_{\wp}}{G_W G_{\wp}}
        m_W m_{\wp} k 
    \Big] \\
   &=
           \frac{1}{P(Z=1)} 
    \frac{1}{P(\zp=1)}
    \E_{\substack{X,\Delta,Z\\
    \xp,\Delta^\prime,
    \zp}}
    \Big[ 
    Z
       \zp 
        k
        +
   \frac{
        \Delta
        \Delta^\prime 
        } 
        {G_W G_{\wp}}
         Z\zp 
        k
        -
        \frac{G_W G_{\wp}}{G_W G_{\wp}}
        Z\zp 
        k
        -
        \frac{\Delta \Delta^\prime
        - G_W G_{\wp}}{G_W G_{\wp}}
        m_W m_{\wp} k 
    \Big] \\
      &=
              \frac{1}{P(Z=1)} 
    \frac{1}{P(\zp=1)}
    \E_{\substack{X,\Delta,Z\\
    \xp,\Delta^\prime,
    \zp}}
    \Big[ 
    Z
       \zp 
        k
        +
   \frac{
        \Delta
        \Delta^\prime 
        -
        G_W G_{\wp}
        } 
        {G_W G_{\wp}}
         Z\zp 
        k
        -
        \frac{\Delta \Delta^\prime
        - G_W G_{\wp}}{G_W G_{\wp}}
        m_W m_{\wp} k 
    \Big] \\
          &=
                  \frac{1}{P(Z=1)} 
    \frac{1}{P(\zp=1)}
    \E_{\substack{X,\Delta,Z\\
    \xp,\Delta^\prime,
    \zp}}
    \Big[ 
    Z
       \zp 
        k
        +
   \frac{
        \Delta
        \Delta^\prime 
        -
        G_W G_{\wp}
        } 
        {G_W G_{\wp}}
        \Big(
         Z\zp 
         -m_W m_{\wp}
         \Big)
        k
    \Big] \\
          &=
                  \frac{1}{P(Z=1)} 
    \frac{1}{P(\zp=1)}
    \E_{\substack{X,\Delta,Z\\
    \xp,\Delta^\prime,
    \zp}}
    \Big[ 
    Z
       \zp 
        k
    \Big] 
        +
         \frac{1}{P(Z=1)} 
    \frac{1}{P(\zp=1)}
        \E_{\substack{X,\Delta,Z\\
    \xp,\Delta^\prime,
    \zp}}
    \Big[ 
   \frac{
        \Delta
        \Delta^\prime 
        -
        G_W G_{\wp}
        } 
        {G_W G_{\wp}}
        \Big(
         Z\zp 
         -m_W m_{\wp}
         \Big)
        k
    \Big] 
\end{align*}
Our estimator equals two terms.
We first show that the first term equals the desired quantity,
and then show the second term equals 0 when either auxiliary model is correct.
\begin{align*}
        \frac{1}{P(Z=1)} 
    \frac{1}{P(\zp=1)}
     \E_{\substack{X,\Delta,Z\\
    \xp,\Delta^\prime,
    \zp}}
    \Big[ 
    Z
       \zp 
        k
    \Big] 
    &=
            \frac{1}{P(Z=1)} 
    \frac{1}{P(\zp=1)}
    \E_{X,\xp}
    \Big[ 
    k \E[Z,\zp|X,\xp]
    \Big] \\
    &=
            \frac{1}{P(Z=1)} 
    \frac{1}{P(\zp=1)}
        \E_{X,\xp}
    \Big[ 
    k 
    P(Z=1,\zp=1
    |X,\xp)
        \Big] \\
        &=
                \frac{1}{P(Z=1)} 
    \frac{1}{P(\zp=1)}
        \int_{X,\xp}
    k 
    P(Z=1,\zp=1
    |X,\xp)
    P(X,\xp)
    dX d\xp \\
    &=
            \frac{1}{P(Z=1)} 
    \frac{1}{P(\zp=1)}
        \int_{X,\xp}
    k 
    P(Z=1|X)
    P(\zp=1|X)
    P(X)P(\xp)
    dX d\xp \\
        &=
            \frac{1}{P(Z=1)} 
    \frac{1}{P(\zp=1)}
        \int_{X,\xp}
    k 
    P(Z=1,X)
    P(\zp=1,\xp)
    dX d\xp \\
    &=
            \frac{1}{P(Z=1)} 
    \frac{1}{P(\zp=1)}
        \int_{X,\xp}
    k 
    P(Z=1,X)
    P(\zp=1,\xp)
    dX d\xp \\
    &=
    \int_{X,\xp}
    k 
    P(X|Z=1)
    P(\xp|\zp=1)
    dX d\xp \\
    &=
    \E_{P(X|Z=1)P(\xp|\zp=1)}
    \Big[
    k 
    \Big] 
\end{align*}
That's the expectation we want missing just the $P(Z=1)$ constants, 
so now we should show the next term is $0$ when either $m$ or $G$ are correct. When $G$ correct:
\begin{align*}
     \E_{\substack{X,\Delta,Z\\
    \xp,\Delta^\prime,
    \zp}}
    \Big[ 
   \frac{
        \Delta
        \Delta^\prime 
        -
        G_W G_{\wp}
        } 
        {G_W G_{\wp}}
        \Big(
         Z\zp 
         -m_W m_{\wp}
         \Big)
        k
    \Big] 
    &=
     \E_{\substack{X,Z\\
    \xp,
    \zp}}
    \Big[ 
   \frac{
    \E[\Delta
        \Delta^\prime 
        |X,\xp,Y,\zp]
        -
        G_W G_{\wp}
        } 
        {G_W G_{\wp}}
        \Big(
         Z\zp 
         -m_W m_{\wp}
         \Big)
        k
    \Big] \\
    &=
     \E_{\substack{X,Z\\
    \xp,
    \zp}}
    \Big[ 
   \frac{
    \E[\Delta
        \Delta^\prime 
        |X,\xp]
        -
        G_W G_{\wp}
        } 
        {G_W G_{\wp}}
        \Big(
         Z\zp 
         -m_W m_{\wp}
         \Big)
        k
    \Big] \\
    &=
     \E_{\substack{X,Z\\
    \xp,
    \zp}}
    \Big[ 
   \frac{
   \E[\Delta|X]
   \E[\Delta^\prime |\xp]
        -
        G_W G_{\wp}
        } 
        {G_W G_{\wp}}
        \Big(
         Z\zp 
         -m_W m_{\wp}
         \Big)
        k
    \Big] \\
    &=
     \E_{\substack{X,Z\\
    \xp,
    \zp}}
    \Big[ 
   \frac{
   G_W G_{\wp}
        -
        G_W G_{\wp}
        } 
        {G_W G_{\wp}}
        \Big(
         Z\zp 
         -m_W m_{\wp}
         \Big)
        k
    \Big]=0
\end{align*}
Likewise, when $m$ correct:
\begin{align*}
     & \E_{\substack{X,\Delta,Z\\
    \xp,\Delta^\prime,
    \zp}}
    \Big[ 
   \frac{
        \Delta
        \Delta^\prime 
        -
        G_W G_{\wp}
        } 
        {G_W G_{\wp}}
        \Big(
         Z\zp 
         -m_W m_{\wp}
         \Big)
        k
    \Big] \\
    &=
      \E_{\substack{X,\Delta\\
    \xp,\Delta^\prime}}
    \Big[ 
   \frac{
        \Delta
        \Delta^\prime 
        -
        G_W G_{\wp}
        } 
        {G_W G_{\wp}}
        \Big(
        \E[Z\zp 
        |X,\xp,
        \Delta,\Delta^\prime]
         -m_W m_{\wp}
         \Big)
        k
    \Big] 
    \\
        &=
      \E_{\substack{X,\Delta\\
    \xp,\Delta^\prime}}
    \Big[ 
   \frac{
        \Delta
        \Delta^\prime 
        -
        G_W G_{\wp}
        } 
        {G_W G_{\wp}}
        \Big(
        \E[Z|X,\Delta] 
        \E[\zp|
        \xp,\Delta^\prime]
         -m_W m_{\wp}
         \Big)
        k
    \Big] 
        \\
        &=
      \E_{\substack{X,\Delta\\
    \xp,\Delta^\prime}}
    \Big[ 
   \frac{
        \Delta
        \Delta^\prime 
        -
        G_W G_{\wp}
        } 
        {G_W G_{\wp}}
        \Big(
        \E[Z|X,\Delta] 
        \E[\zp|
        \xp,\Delta^\prime]
        -
        \E[Z|X,\Delta=1] 
        \E[\zp|
        \xp,\Delta^\prime=1]
         \Big)
        k
    \Big] \\
            &=
      \E_{\substack{X,\Delta\\
    \xp,\Delta^\prime}}
    \Big[ 
   \frac{
        \Delta
        \Delta^\prime 
        -
        G_W G_{\wp}
        } 
        {G_W G_{\wp}}
        \Big(
        \E[Z|X] 
        \E[\zp|
        \xp]
        -
        \E[Z|X] 
        \E[\zp|
        \xp]
         \Big)
        k
    \Big] =0
\end{align*}

The proof for the other two terms is analogous but with using $\overline{Z}=(1-Z)$ instead of $Z$ and $\overline{m}=1-m$ when conditioning on $Z=0$.

\section{kernel mmd between joint and product of marginals
\label{appsec:mmdcontinuous}} 

\paragraph{Continuous nuisances.} In this work we study binary nuisance. We can instead measure the \gls{mmd} between joint $p(h_X,Z)$ and product of marginals $p(h_X)P(Z)$, which allows for continuous nuisance. 

The above formulation of \gls{mmd} between
$h_X|Z=1$ and  $h_X|Z=0$ relied on optimizing with respect to $h$ only: $P(Z)$ is constant in the optimization so the distance between conditionals specifies the distance between the product of marginals and joint and thus the dependence. However, considering the more general case of \gls{mmd} between $P(h_X,Z)$ and $P(h_X)P(Z)$ has the advantage that is not necessary to consider a finite set of conditioning values for $Z$. That means the \gls{mmd} can be extended to continuous nuisance $Z$. 
Let $X::Z$ denote the concatenation of $X$ and $Z$.
The more general formulation is:
\begin{align*}
    \mathbb{E}_{
        \substack{(X,Z) \sim P(X,Z)\\
                (\xp,\zp) \sim P(X,Z)}
    }&\Bigg[k \Big(X \cat \cat Z,\xp \cat \cat \zp \Big) \Bigg]
    +
        \mathbb{E}_{
        \substack{(X,Z) \sim P(X)P(Z)\\
                (\xp,\zp) \sim P(\xp)P(\zp)}
    }\Bigg[k \Big(X \cat \cat Z,\xp \cat \cat \zp \Big) \Bigg]\\
   & - 2
       \mathbb{E}_{
        \substack{(X,Z) \sim P(X,Z)\\
                (\xp,\zp) \sim P(\xp)P(\zp)}
    }\Bigg[k \Big(X \cat \cat Z,\xp \cat \cat \zp \Big) \Bigg]
\end{align*}
This leads to the following estimator:
\begin{align*}
\mathbb{E}_{
        \substack{P(X,Z)\\
                P(\xp,\zp)}
    }\Bigg[k \Big(X \cat \cat Z,\xp \cat \cat \zp \Big) \Bigg]
    &=
    \mathbb{E}
    \Bigg[
    \frac{\Delta \Delta^\prime k(X \cat \cat  Z,\xp \cat \cat \zp)}{G_{W \wp}}
    - 
    \frac{\Delta \Delta^\prime - G_{W \wp}}{G_{W \wp}}\E[k|X,\xp]
\Bigg]
\end{align*}
and
\begin{align*} 
  \mathbb{E}_{
        \substack{P(X)P(Z)\\
        P(\xp)P(\zp)}
    }&\Bigg[k \Big(X \cat \cat Z, \xp \cat \cat \zp \Big) \Bigg] \\
    &= 
  \mathbb{E}_{
        \substack{ P(X_1)P(X_2,Z_2)\\
     P(X_3)P(X_4,Z_4)}
    }\Bigg[k \Big(X_1 \cat \cat Z_2,X_3 \cat \cat Z_4 \Big) \Bigg]\\
&=
    \mathbb{E}
    \Bigg[
    \frac{\Delta \Delta^\prime k (X_1 \cat \cat Z_2,X_3 \cat \cat Z_4)}{G_{X_1 X_3}}
    - 
    \frac{\Delta \Delta^\prime - G_{X_1 X_3}}{G_{X_1 X_3}}\E[k (X_1 \cat \cat Z_2,X_3 \cat \cat Z_4)|X_1,X_3] \Bigg]
    \end{align*}
    and 
    \begin{align*} 
  \mathbb{E}_{
        \substack{P(X,Z)\\
        P(\xp)P(\zp)}
    }&\Bigg[k \Big(X \cat \cat Z, \xp \cat \cat \zp \Big) \Bigg] \\&= 
  \mathbb{E}_{
        \substack{ P(X_1,Z_1)\\
     P(X_2)P(X_3,Z_3)}
    }\Bigg[k \Big(X_1 \cat \cat Z_1,X_2 \cat \cat Z_3 \Big) \Bigg]\\
    &=
    \mathbb{E}
    \Bigg[
    \frac{\Delta \Delta^\prime k(X_1 \cat \cat Z_1,X_2 \cat \cat Z_3)}{G_{X_1 X_3}}
    - 
    \frac{\Delta \Delta^\prime - G_{X_1 X_3}}{G_{X_1 X_3}}\E[k(X_1 \cat \cat Z_1,X_2 \cat \cat Z_3)|X_1,X_3] \Bigg] 
\end{align*}
The challenging part of applying this estimator is that now instead of one function $m_W$ we have three functions,
each of which estimates the mean of $k$ under a different sampling distribution. Moreover, these conditional expectations depend on the current representation $h_X$. This means they must be updated each time $h$ changes.

\end{document}